\documentclass[runningheads]{llncs}
\usepackage{graphicx}

\usepackage{tikz}
\usepackage{array}
\newcolumntype{C}[1]{>{\centering\arraybackslash}p{#1}}
\usepackage{comment}
\usepackage{amsmath,amssymb} 
\usepackage{color}
\usepackage{booktabs}
\usepackage{lipsum}  
\usepackage{pifont}
\usepackage[accsupp]{axessibility}  
\usepackage{subcaption}

\usepackage[colorlinks]{hyperref}
\usepackage[capitalize]{cleveref}

\begin{document}

\pagestyle{headings}
\mainmatter
\def\ECCVSubNumber{5285}  

\title{Approximate Differentiable Rendering \\with Algebraic Surfaces} 

\titlerunning{Approximate Differentiable Rendering with Algebraic Surfaces }
%
\author{Leonid Keselman \and
Martial Hebert}
\authorrunning{ Leonid Keselman and Martial Hebert}
%
\institute{Carnegie Mellon University, Pittsburgh PA, USA\\
\email{\{lkeselma,hebert\}@cs.cmu.edu}}
\maketitle

\begin{abstract}
Differentiable renderers provide a direct mathematical link between an object's 3D representation and images of that object. In this work, we develop an approximate differentiable renderer for a compact, interpretable representation, which we call Fuzzy Metaballs. Our approximate renderer focuses on rendering shapes via depth maps and silhouettes. It sacrifices fidelity for utility, producing fast runtimes and high-quality gradient information that can be used to solve vision tasks. Compared to mesh-based differentiable renderers, our method has forward passes that are 5x faster and backwards passes that are 30x faster. The depth maps and silhouette images generated by our method are smooth and defined everywhere. In our evaluation of differentiable renderers for pose estimation, we show that our method is the only one comparable to classic techniques. In shape from silhouette, our method performs well using only gradient descent and a per-pixel loss, without any surrogate losses or regularization. These reconstructions work well even on natural video sequences with segmentation artifacts. \\ Project page: \url{https://leonidk.github.io/fuzzy-metaballs}
\keywords{differentiable rendering, metaballs, implicit surfaces, pose estimation, shape from silhouette, gaussian mixture models}
\end{abstract}

\section{Introduction}\label{sec:intro}
Rendering can be seen as the inverse of computer vision: turning 3D scene descriptions into plausible images. There are countless classic rendering methods, spanning from the extremely fast (as used in video games) to the extremely realistic (as used in film and animation). Common to all of these methods is that the rendering process for opaque objects is discontinuous; rays that hit no objects have no relationship to scene geometry and  when intersections do occur, they typically only interact with the front-most component of geometry.  

Differentiable Rendering is a recent development, designing techniques (often sub-gradients) that enable a more direct mathematical relationship between an image and the scene or camera parameters that generated it. The easy access to derivatives allows for statistical optimization and natural integration with gradient-based learning techniques. There exist several recent differentiable renderers which produce images comparable in fidelity to classic, non-differentiable, photorealistic rendering methods~\cite{Laine2020diffrast,nerf20,Mitsuba2019diff,Zhang:2020:PSDR}. 

\begin{figure*}[th]
  \centering
   \includegraphics[width=0.95\linewidth]{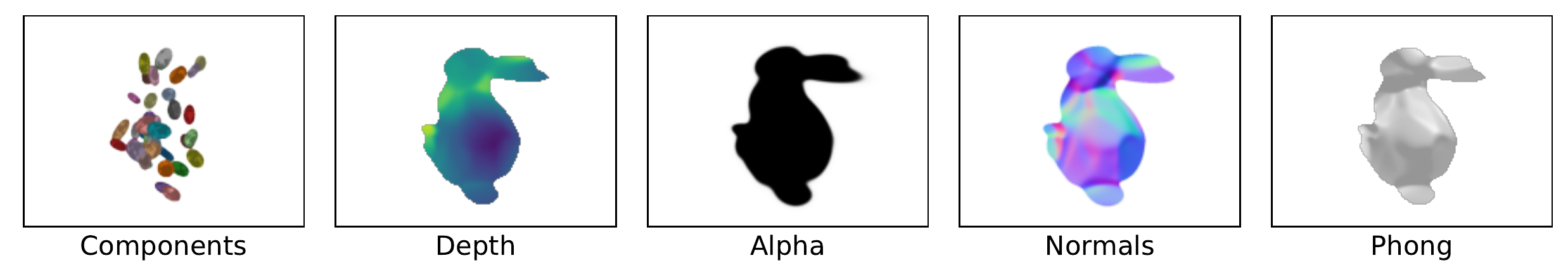}
   \caption{\textbf{Our differentiable renderer producing images of Stanford bunny}, using a representation with 400 parameters. From left to right: the 40 components at one standard deviation, followed by our differentiable renderer generating depth, alpha, surface normals and a shaded image.}
   \label{fig:onecol}
\end{figure*}

Our paper presents a different approach: a differentiable renderer focused on utility for computer vision tasks. We are interested in the quality and computability of gradients, not on matching the exact image formation task.  We willingly sacrifice fidelity for computational simplicity (and hence speed). Our method focuses on a rendering-like process for shapes which generates good gradients that rapidly lead to viable solutions for classic vision problems. Other methods may produce more pleasing images, but we care about the quality of our local minima and our ability to easily find those minima. Our experiments show how, compared to classic methods, differentiable renderers can be used to solve classic vision problems using only gradient descent, enabling a high degree of robustness to noise such as under-segmented masks or depth sensor artifacts.

Our approach is built on a specific 3D representation. Existing representations often have undesirable properties for rendering or optimization. Point clouds require splatting or calculating precise point sizes~\cite{DSS_points}. Meshes explicitly represent the object surface, making changes of genus difficult. Other representations require optimization or numerical estimation of ray-shape intersections~\cite{10.1145/357306.357310,nerf20}. Our proposed method is formulated with independent rays, represents object surfaces implicitly and computes ray termination in closed form. 


Most existing differentiable renders focus on GPU performance. However, GPUs are not always available. Many robotics platforms do not have a GPU~\cite{6290694} or find it occupied running object detection~\cite{8708251}, optical flow~\cite{teedraft20} or a SLAM method~\cite{miller2020tunnel}. While a single method may claim to be real-time on a dedicated GPU~\cite{teed2021droidslam}, an autonomous system requires a sharing of resources. To run in parallel with the countless GPU-friendly techniques of today, CPU-friendly methods are desirable. Thus, while our method is implemented in JAX~\cite{jax2018github}, supporting CPU and GPU backends, our focus is typically on CPU runtimes. 

Lastly, in the era of deep learning, techniques which support gradient-based optimization are desirable. Since our objects have an explicit algebraic form, gradients are simple and easy to compute. Importantly, every pixel has a non-zero (if very slight) relationship with each piece of geometry in the scene (even those behind the camera!). This allows for gradient flow (up to machine precision), even when objects start far from their initialization. While this can also true of large over-parmaterized implicit surfaces (such as NeRF~\cite{nerf20}), our representation is extremely compact and each parameter has approximate geometric meaning. 

\section{Related Work}\label{sec:related}
Early work in 3D shape representation focused on building volumes from partial observations~\cite{10.5555/905981} but most modern methods instead focus on surface representation. Meshes, point clouds and surfels~\cite{10.1145/344779.344936}  focus on representing the exterior of an object. In contrast, our method works by representing volumes, and obtaining surface samples is implicit; similar to recent work on implicit neural surfaces~\cite{nerf20}.

In using low-fidelity representations, our work is hardly unique. Often learning-based methods settle for pseudorendering~\cite{lin2018learning} or even treating images as layers of planar objects~\cite{tucker2020single}. Settling for low fidelity contrasts sharply with a wide array of differentiable renderers focused on accurate light transport, which are slower but can simulate subtle phenomena~\cite{bangaru2020warpedsampling,Zhang:2020:PSDR}. High-quality results can also be obtained by using learning methods and dense voxel grids~\cite{Lombardi:2019}. 

Differentiable Rendering has many recent works. OpenDR~\cite{OpenDR}  demonstrated pose updates for meshes representing humans. Neural Mesh Renderer~\cite{kato2017neural} developed approximate gradients and used a differentiable renderer for a wide array of tasks. SoftRasterizer~\cite{liu2019soft} developed a subgradient function for meshes with greatly improved gradient quality. Modular Primitives~\cite{Laine2020diffrast} demonstrated fast, GPU-based differentiable rendering for meshes with texture mapping. Differentiable Surface Splatting~\cite{DSS_points} developed a differentiable renderer for point clouds by building upon existing rendering techniques~\cite{10.1145/383259.383300}. Conversion of point clouds to volumes is also differentiable~\cite{insafutdinov2018unsupervised}. Pulsar~\cite{lassner2020pulsar} uses spheres as the primary primitive and focuses on GPU performance.  PyTorch3D~\cite{ravi2020pytorch3d} implements several of these techniques for mesh and point cloud rendering. Some methods exploit sampling to be generic across object representation~\cite{cole2021differentiable}. Many methods integrate with neural networks for specific tasks, such as obtaining better descriptors~\cite{li2020endtoend} or predicting 3D object shape from a single images~\cite{chen2019learning,8237663}. 

The use of an algebraic surface representation, which came to be known as \textit{metaballs} can be attributed to Blinn~\cite{10.1145/357306.357310}. These algebraic representations were well studied in the 1980s and 1990s. These include the development of ray-tracing approximations ~\cite{Heckbert85funwith,Wyvill1986,raytraceSoft} and building metaball representations of depth images~\cite{10.1145/122718.122743}. Non-differentiable rendering metaballs has many methods, involving splatting~\cite{ALD2006PSIROPSD}, data structures~\cite{gourmel:hal-01516266,Szcsi2012RealTimeMR} or even a neural network~\cite{horvath-2018-ism}.

Metaballs, especially in our treatment of them, are related to the use of Gaussian Mixture Models (GMMs) for surface representation. Our method could be considered a ~\textit{differentiable renderer for GMMs}. Gaussian Mixture Models as a shape representation has some appeal to roboticists~\cite{8586902,wenniegmm}. Methods developed to render GMMs include search-based methods~\cite{shankar20mrfmap} and projection for occupancy maps~\cite{8586902}. Projection methods for GMMs have also found application in robot pose estimation~\cite{9126150}. In the vision community, GMMs have been studied as a shape representation~\cite{Eckart2016} and  used for pose estimation~\cite{Eckart2018,Eckart2015}. In the visual learning space, GMMs~\cite{hertz2019pointgmm}, or their approximations~\cite{genova2020local} have also been used. 

Concurrent work also uses Gaussians for rendering. VoGE~\cite{wang2022voge} uses existing volume rendering techniques~\cite{468400,nerf20}. Others use a DGT to build screen-space Gaussians for point clouds~\cite{sampling2022ECCV}. In contrast, our contribution is the development of an approximate differentiable renderer that produces fast \& robust results. 

\section{Fuzzy Metaballs}\label{sec:method}
Our proposed object representation, dubbed Fuzzy Metaballs, is an algebraic, implicit surface representation. Implicit surfaces are an object representations where the surface is represented as
\begin{equation}\label{eq:implicit}
    F(x,y,z) = 0.
\end{equation}

While some methods parameterize $F$ with neural networks~\cite{nerf20},  Blinn's algebraic surfaces~\cite{10.1145/357306.357310}, also known as  blobby models or metaballs, are defined by
\begin{equation}\label{eq:summ_blob}
    F(x,y,z) = \sum_i \lambda_i P(x,y,z) - T,
\end{equation}
where $P(x,y,z)$ is some geometric component and $T$ is a threshold for sufficient density. While Blinn used isotropic Gaussians (hence balls), in our case, we use general multidimensional Gaussians that form ellipsoids:
\begin{equation}\label{eq:component}
    P(\vec{x}) = |\Sigma|^{-\frac{1}{2}} \exp\left(-\frac{1}{2} (\vec{x}-\mu)^T \Sigma^{-1} (\vec{x}-\mu)\right).
\end{equation}

In contrast to classic metaballs, we relax the restriction on $T$ being a hard threshold set by the user and instead develop a ray-tracing formulation for Gaussians which implicitly defines the surface; hence \textit{fuzzy} metaballs. To achieve this, we develop two components: a way of defining intersections between Gaussians and rays (\cref{sec:intersect}), and a way of combining intersections across all Gaussians (\cref{sec:blend}). In our definition, all rays always intersect all Gaussians, leading to smooth gradients. The \textit{fuzzy} surface locations are not viewpoint invariant. 

Our implementation is in JAX~\cite{jax2018github}, enabling CPU and GPU acceleration as well as automatic backpropogation. The rendering function that takes camera pose, camera rays and geometry is 60 lines of code. To enable constraint-free backpropogation, we parameterize $\Sigma^{-1}$ with its Cholesky decomposition: a lower triangular matrix with positive diagonal components. We ensure that the diagonal elements are positive and at least $10^{-6}$. The determinant is directly computed from a product of the diagonal of $L$.  When analyzing ray intersections, one can omit the $|\Sigma|^{-\frac{1}{2}}$ term as maximizing requires only the quadratic form. For example, $\vec{x}$ is replaced with a ray intersection of $\vec{v} t$ with $\vec{v} \in \mathbf{R}^3$ and $t \in \mathbf{R}$: 
\begin{equation}\label{eq:quad_form}
    s(v t) = (v t -\mu)^T \Sigma^{-1} (v t-\mu),
\end{equation}
giving a Mahlanobis distnance~\cite{Mahalanobis1936OnTG} that is invariant to object scale and allows us to use constant hyperparameters, irrespective of object distance. Using probabilities would be scale-sensitive as equivalent Gaussians that are further are also larger and would have smaller likelihoods at the same points.

To produce an alpha-mask, we simply have two hyperparameters for scale and offset and use a standard sigmoid function:
\begin{equation}
    \alpha = \sigma\left(\beta_4  \left[\sum_i \lambda_i \exp(-\frac{1}{2} s(v t) )\right] + \beta_5\right).
\end{equation}

\section{Approximate Differentiable Rendering}
Instead of using existing rendering methods, we develop an approximate renderer that produces smooth, high-quality gradients. While inexact, our formulation enables fast and robust differentiable rendering usable in an analysis by synthesis pipeline~\cite{doi:10.1121/1.1908556}. We split the process into two steps: intersecting each component independently in \cref{sec:intersect} and combining those results smoothly in \cref{sec:blend}. 

\subsection{Intersecting Gaussians}\label{sec:intersect}
What does it mean to have a particular intersection of a ray with a Gaussian? We propose three methods. The \textit{linear} method is where the ray intersects the Gaussian at the point of highest probability. Maximizing~\cref{eq:quad_form} is solved by

\begin{equation}\label{eq:linear}
    t = \frac{\mu^T \Sigma^{-1} v}{v^T \Sigma^{-1} v}.
\end{equation}

An alternative view is a volume model,  intersecting at the maximum magnitude of the gradient of the Gaussian:
\begin{equation}\label{eq:cubic}
    ||\nabla p(t v)||^2 = P(t v)^2 = (t v-\mu)^T\Sigma^{-1} \Sigma^{-1}(t v-\mu).
\end{equation}
Obtaining the gradient of ~\cref{eq:cubic} and setting it equal to zero leads to a cubic equation, hence the \textit{cubic} method. Defining $ m = \Sigma \mu$ and $r = \Sigma v$ leads to:
\begin{gather*}
0 = - t^3( r^T r) (v^T r) \\ 
    +  t^2 \left[ ( m^T r + r^T m) (v^T r) + ( r^T r) (v^T m) \right] \\
    -  t \left[ (m^T m) (v^T r) + ( m^T r + r^T m)  (v^T m) - r^T r \right] \\
    +   (m^T m) (v^T m) -  r^T m.
\end{gather*}

While standard formulas exist for the cubic, the higher order polynomial all-but-ensures that numerical issues will arise. We implement a numerically stable solver for the cubic~\cite{4178164}. However, even the numerically stable version produces problematic pixels in 32bit floating point. Errors at a rate of about 1 in 1,000 produce NaNs and make backpropagation impossible. 


The \textit{quadratic} method approximates the cubic by intersecting the Gaussian at the one standard deviation ellipsoid. Clipping the inside of square roots to be non-negative leads to reasonable results when the ray misses the ellipsoid. 
$$t^2 v^T \Sigma^{-1} v - 2 t v^T \Sigma^{-1} \mu +  \mu^T\Sigma^{-1} \mu = 1 $$
$$ a = v^T \Sigma^{-1} v \qquad b = -2 v^T \Sigma^{-1} \mu \qquad c = \mu^T\Sigma^{-1} \mu -1$$

\begin{figure*}
  \centering
   \includegraphics[width=0.9\linewidth]{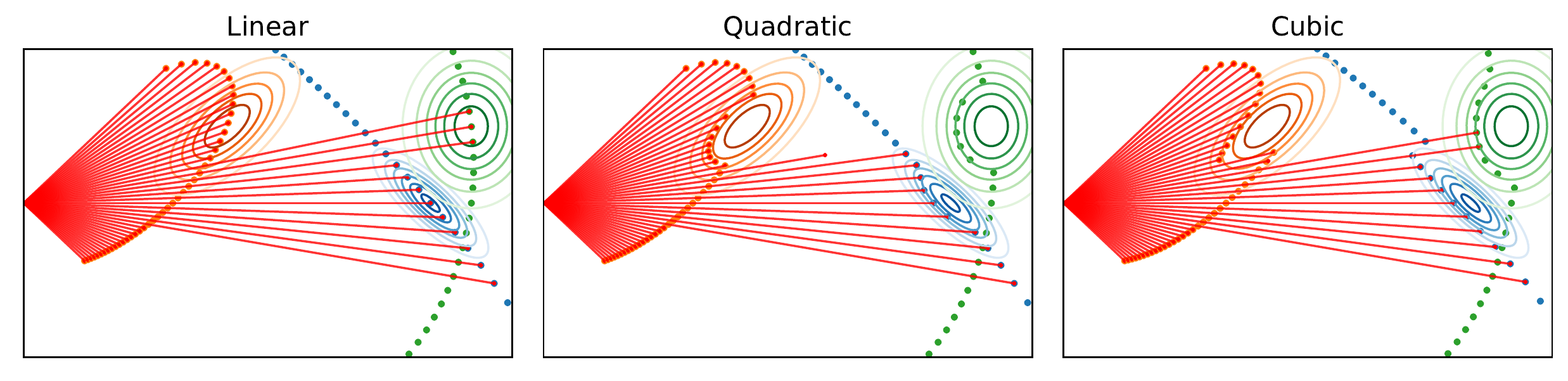}
    \caption{Two dimensional version of our approximate renderer with camera rays cast from the center left. Three components are shown by their contour maps and their intersections with dots. The blended results are shown with red rays. }
    \label{fig:2dray}
\end{figure*}
\begin{figure}[tb]
  \centering
   \includegraphics[width=0.9\linewidth]{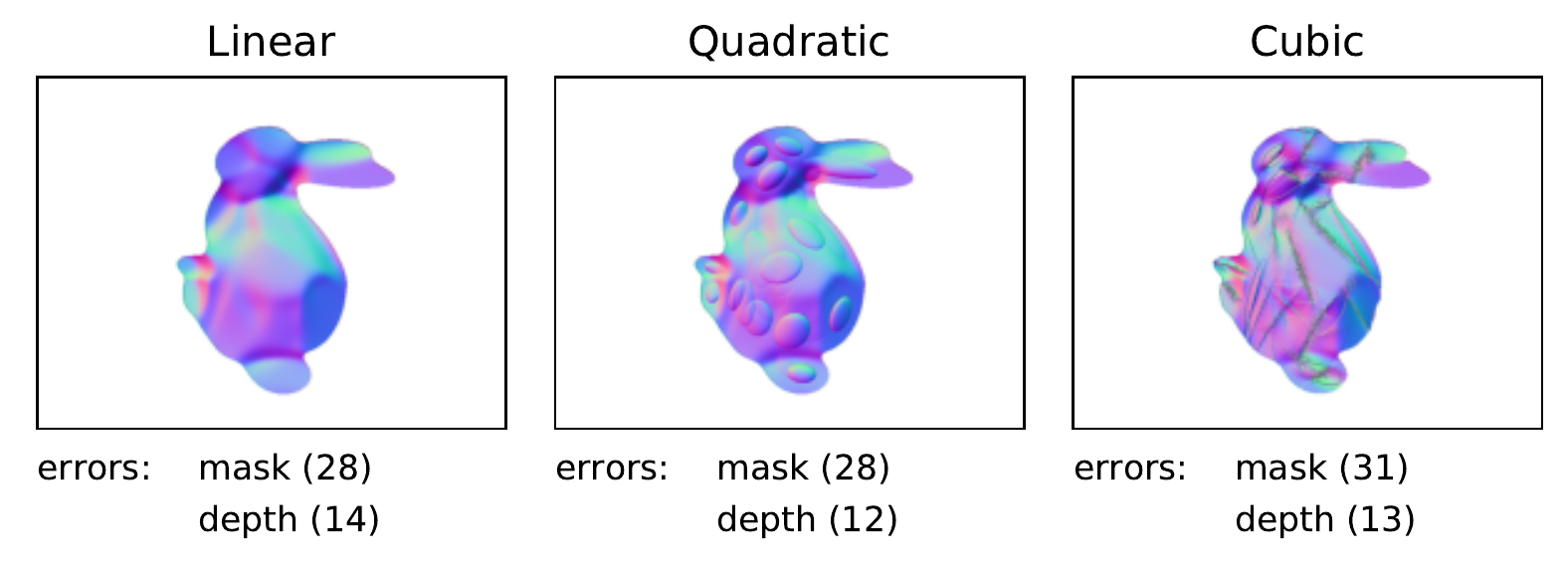}
    \caption{Visual examples of  normal maps from different methods of ray intersection, along with the respective mask and depth errors. See ~\cref{sec:intersect} for details}
    \label{fig:3dray}
\end{figure}

\Cref{fig:2dray,fig:3dray} illustrate all three methods. The linear method produces smooth surfaces and the quadratic surface shows the individual ellipsoids protruding from the surface of the object and the cubic shows artifacts. 

In 3D evaluation on objects, for a forward pass, the linear method is the fastest, the quadratic method takes 50\% longer and the cubic method takes twice as long as the linear method. The quadratic method has the lowest errors in depth and mask errors. However, due to its stability, in all evaluation outside this section, we use the linear method. 

\subsection{Blending intersections}\label{sec:blend}
We present a particular solution to the hidden-surface problem~\cite{sutherland1974hidden}. Our method is related to prior work on \textit{Order Independent Transparency (OIT)} ~\cite{SA09,McGuire2013Transparency} but extended to 3D objects with opaque surfaces. We combine each pixel's ray-Gaussian intersections with a weighted average
\begin{equation}\label{eq:blend1}
    t_f = \frac{1}{\sum_i w_i} \sum_i w_i t_i.
\end{equation}

The weights are an exponential function with two hyperparameters $\beta_1$ and $\beta_2$ balancing high-quality hits versus hits closer closer to the camera:
\begin{equation}\label{eq:blend2}
    w_i = \exp\left(\beta_1 s(v t_i) h(t_i) -\frac{\beta_2}{\eta} t_i\right).
\end{equation}
We include a term ($\eta$) for the rough scale of the object. This, along with use of ~\cref{eq:quad_form} allows our rendering to be invariant to object scale. We also include an extra term to down-weight results of intersections behind the camera with a simple sigmoid function:
\begin{equation}
    h(t) = \sigma\left(\frac{\beta_3}{\eta} t\right).
\end{equation}
Our blending solution requires only $O(N)$ evaluations of Gaussians for each ray. 
\subsection{Obtaining Fuzzy Metaballs}\label{sec:fitting}
A representation can be limited in utility by how easily one can convert to it. We propose that, unlike classic Metaballs, Fuzzy Metaballs have reasonably straightforward methods for conversion from other formats. 

Since we've developed a differentiable renderer, one can optimize a Fuzzy Metaball representation from a set of images. One could use several different losses, but experiments with silhouettes are described in \cref{sec:recon}.

If one has a mesh, the mathematical relationship of Fuzzy Metaballs and Gaussian Mixture Models can be exploited by fitting a GMM with Expectation-Maximization~\cite{Dempster1977}. With Fuzzy Metaballs being between a surface and volume representation, there are two forms of GMM one could fit. The first is a surface GMM (sGMM) as used by many authors~\cite{Eckart2016,8781611,wenniegmm}, where a GMM is fit to points sampled from the surface of the object. The second is to build a volumetric GMM (vGMM). To build a vGMM, one takes a watertight mesh~\cite{SGP:SGP06:061-070}, and samples points from the interior of the object. Fitting a GMM to these interior points is what we call a volumetric GMM. Both representations can then further be optimized using the differentiable renderer. Our experiments show that both forms of GMM initialization work well, but we use vGMMs in our experiments. 

Extraction is also straightforward. Point clouds can easily be sampled from our proper probability distributions.  Extracting a mesh is possible by  running marching cubes~\cite{lorensen1987marching} with an optimized iso-surface level. Details in \cref{sec:marching}.

\begin{table*}[tb]
  \centering
    \caption{\textbf{Runtimes in milliseconds} with $\mu \pm \sigma$ for rendering images and performing gradient updates in pose estimation with comparable fidelity (\cref{sec:fid}). CPU performance may be a fairer comparison as our method is 60 lines of JAX~\cite{jax2018github} code and lacks a custom CUDA kernel. CUDA numbers use 160 x 120 images on a Quadro P2000, while CPU use 80 x 60 images on an i5-7287U. }
  \begin{tabular}{@{}p{0.3\textwidth}C{0.15\textwidth}C{0.15\textwidth}|C{0.15\textwidth}C{0.15\textwidth}@{}}
    \toprule
    \textbf{Method} &  \textbf{Forward CUDA} & \textbf{Backwards CUDA} & \textbf{Forward CPU} & \textbf{Backwards CPU}  \\
       \midrule
    Point Cloud~\cite{ravi2020pytorch3d}  & $ 12.1 \pm 0.5$ &  $ 23.4 \pm 0.5$ &  $ 18.0 \pm 1.0$ &  $ 23.8 \pm 4.0$ \\
    Pulsar~\cite{lassner2020pulsar}    & $ 7.8 \pm 0.3$ &  $ 11.2 \pm 0.4$& $ 16.4 \pm 1.4$ &  $ 63.6 \pm 7.9$ \\
    SoftRas Mesh~\cite{liu2019soft,ravi2020pytorch3d} & $ 17.0 \pm 0.4$ &  $ 27.2 \pm 0.5$  & $ 21.5 \pm 2.0$ &  $ 384.7 \pm 93.8$ \\
    \midrule
    Fuzzy Metaballs &   \textbf{3.0} $\pm$ \textbf{0.2} &  \textbf{9.6} $\pm$ \textbf{0.5} & \textbf{3.0} $\pm$ \textbf{0.15} &  \textbf{13.2} $\pm$ \textbf{1.4} \\
    \bottomrule
  \end{tabular}

  \label{tab:runtime}

\end{table*}

\section{Data}\label{sec:data}
We use ten models for evaluation: five from the Stanford Model Repository ~\cite{levoy2005stanford} (arma, buddha, dragon, lucy, bunny), three from Thingi10K~\cite{DBLP:journals/corr/ZhouJ16} (gear, eiffel, rebel) and two from prior rendering literature (yoga, plane). All ten are used for reconstruction, and seven are used for pose estimation. We selected objects with different scales, genus, and variety in features. We choose 40 component FMs based on prior literature suggesting 20 to 60 GMMs for object representation~\cite{Eckart2018}.



\begin{figure}[tbh]
  \centering
   \includegraphics[width=1.0\linewidth]{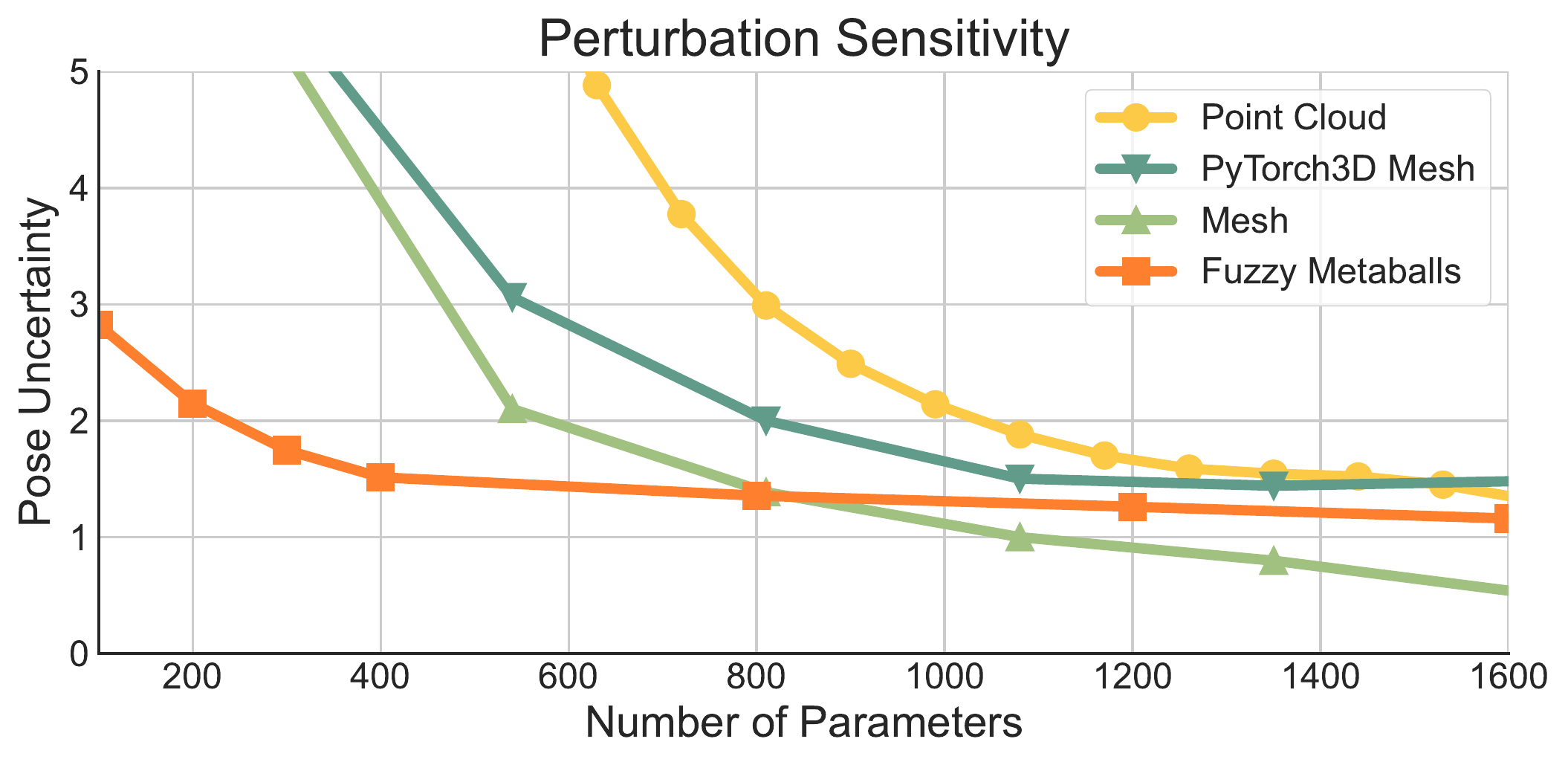}
    \caption{\textbf{Perturbation sensitivity} is the average error in pose when registration is performed with ground truth pose as initialization. See \cref{sec:fid} for details. The underlying ground truth is a decimated mesh, so only the mesh representation approaches exactly zero error while other asymptote at a higher mark. }
    \label{fig:fidelity_comp}
\end{figure}

\section{Comparing Representations}\label{sec:fid}
Fairly comparing object representations requires some notion of what to hold constant. As the parameter counts of each representation increase, so do their representational ability. It would be unfair to compare a million point point-cloud against a 100 face triangle mesh. Since our goal is utility in vision tasks, our definition of fidelity will also be task-centric. 

In this case,  our metric of fidelity will be a representation's \textit{perturbation sensitivity}. We define this as the pose error obtained when optimizing an object's camera pose given a depth map, when the optimization process was initialized with ground truth camera pose. The given depth map is of the full representation object, but the methods are evaluated using lower fidelity versions, leading to perturbations of optimal pose and our fidelity metric. Pose errors are reported using the geometric mean of rotation error and translation error. 

Results of our fidelity experiments can be seen in ~\cref{fig:fidelity_comp}. We evaluate point clouds and meshes using a standard Iterative Closest Point (ICP) method~\cite{Zhou2018}, with the point clouds randomly subsampled and the meshes undergoing decimation~\cite{Garland1997}. We also use PyTorch3D~\cite{ravi2020pytorch3d}, a differential mesh renderer, and obtain its perturbation curve. These experiments are conditional on an experimental setup and methods used, and thus these results may change under different conditions.

In our experiments, a 40 component Fuzzy Metaball (the size we throughout across this paper) produces a pose uncertainty equivalent to a 470 point point cloud (roughly triple the parameter count of a fuzzy metaball) and 85 vertex, 170 triangle mesh (roughly twice the parameter count). These are the sizes use throughout the rest of the paper, in our attempt to keep comparisons fair.

\section{Experiments}\label{sec:results}
For comparison against other Differentiable Renderers, we use the methods implemented in PyTorch3D~\cite{ravi2020pytorch3d}, which has a wide variety of techniques with well-optimized routines. The mesh rendering method is an implementation of the \textit{SoftRasterizer}~\cite{liu2019soft}. For point clouds, PyTorch3D cites \textit{Direct Surface Splatting}~\cite{DSS_points}, while also implementing the recent \textit{Pulsar}~\cite{lassner2020pulsar}.

With the fidelity of different object representations normalized out (\cref{sec:fid}), we can compare the runtime performance in a fair way, with times shown in in~\cref{tab:runtime}. On the CPU, where comparisons are more equal (due to lacking a custom CUDA kernel), our renderer is 5 times faster for a forward pass, and significantly faster (30x) for a backwards pass compared to the mesh rendering methods. The point cloud renderer is more comparable in runtime to ours but need a pre-specified point size, often producing images with lots of holes (when points are too small) or a poor silhouette (when points are too big). 

 To the demonstrate the ability our differentiable renderer to solve classic computer vision tasks, we look at pose estimation (\cref{sec:pose}) and 3D reconstruction from silhouettes (\cref{sec:recon}). Our renderer is a function that takes camera pose and geometry, and produces images. It seems natural to take images and evaluate how well either camera pose or geometry can be reconstructed, when the other is given.  All five hyperparameters for our rendering algorithm ($\beta_{1,2,3,4,5}$) were held constant throughout all experiments. 

Since pose estimation and shape from silhouette (SFS) are classic computer vision problems, there are countless methods for both tasks. We do not claim to be the best solution to these problems, as there are many methods specifically designed for these tasks under a variety of conditions. Instead, we seek to demonstrate how our approximate differentiable renderer is comparable in quality to typical solutions, using only gradient descent, without any regularization.

\begin{table}[th!]
\centering
\caption{\textbf{Pose Estimation Results}. Pose Errors are reported with a geometric mean of rotation and translation error. The reported numbers are $ \text{mean} \pm \text{IQR}$. We report results clean data and data with simulated depth and silhouette noise. }
\begin{tabular}{@{}p{0.36\textwidth}rC{0.26\textwidth}C{0.19\textwidth}@{}}
\toprule
 & \textbf{Parameters} & \textbf{Noise-Free Error} & \textbf{Noisy Error} \\ \midrule
Initialization &  & 20.2 $\pm$ 18 & 20.2 $\pm$ 18 \\ \midrule
Pulsar~\cite{lassner2020pulsar} & 1,200 & 20.2  $\pm$ 18 & 20.2  $\pm$ 18 \\
Point Cloud~\cite{ravi2020pytorch3d} & 1,200 & 18.5  $\pm$ 16 & 18.4  $\pm$ 16 \\
SoftRas Mesh~\cite{liu2019soft} & 750 & 14.9 $\pm$ 15 & 17.0 $\pm$ 17 \\ \midrule
Equal Fidelity ICP (Plane)~\cite{Zhou2018} & 1,200 & 10.8 $\pm$ 12 & 8.2 $\pm$ 3.3 \\
Equal Fidelity ICP (Point)~\cite{Zhou2018} & 1,200 & 7.6 $\pm$ 9.9 & 8.7 $\pm$ 6.6 \\ \midrule
High Fidelity ICP  (Plane)~\cite{Zhou2018} & 120,000 & 8.2 $\pm$ 0.8 & 8.0 $\pm$ 3.6 \\
High Fidelity ICP  (Point)~\cite{Zhou2018} & 120,000 & 6.2 $\pm$ 3.7 & 6.8 $\pm$ 3.3 \\ \midrule
Fuzzy Metaballs & 400 & \textbf{4.0} $\pm$ \textbf{1.5} & \textbf{4.2}  $\pm$ \textbf{2.1} \\
\bottomrule
\end{tabular}
\label{tab:pose_est_results}
\end{table}

\subsection{Pose Estimation}\label{sec:pose}
Many differential renderers show qualitative results of pose estimation~\cite{liu2019soft,DSS_points}. We instead perform quantitative results over our library of models rendered from random viewpoints. Methods are given a perturbed camera pose ($\pm 45^{\circ}$ rotation and a random translation up to 50\% of model scale) and the ground truth depth image from the original pose. The methods are evaluated by their ability to recover the original pose from minimizing image-based errors. The resulting pose is evaluated for rotation error and translation error. We quantify the score for a model as the geometric mean of the two errors. All methods are tested on the same random viewpoints and with the same random perturbations. 

For Fuzzy Metaballs, we establish projective correspondence~\cite{efficient_icp} and optimize silhouette cross-entropy loss averaged over all pixels:
\begin{equation}\label{eq:cross_entropy}
    CE(\alpha,\hat{\alpha}) = \alpha \log(\hat{\alpha}) + (1-\alpha) \log(1-\hat{\alpha}).
\end{equation}
Estimated alpha is clipped to $[10^{-6},1-10^{-6}]$ to avoid infinite error. We also evaluate with an additional depth loss of $MSE(z,\hat{z})$ where $|z|$ normalizes the errors to be invariant to object scale and comparable in magnitude to $CE(\alpha,\hat{\alpha})$. 
\begin{equation}\label{eq:mse}
    MSE(z,\hat{z}) = \left|\left|\frac{(z-\hat{z})}{|z|} \right|\right|_2
\end{equation}

There is a subtle caveat in the gradients of Fuzzy Metaballs. The gradient of the translation scales by the inverse of model scale.. We correct for this by scaling the gradients by $\eta^2$. Alternatively one could scale the input data to always be of some canonical scale~\cite{6751291}. To maintain scale invariance, we limit our use of adaptive learning rate methods to SGD with Momentum. 

We provide point cloud ICP results for point-to-point and point-to-plane methods~\cite{efficient_icp} as implemented by Open3D~\cite{Zhou2018}. For the differentiable rendering experiments, we use PyTorch3D~\cite{ravi2020pytorch3d} and tune its settings (\cref{sec:tune_softras}). All differentiable rendering methods use the same loss,  learning rate decay criteria and are run until the loss stops reliably decreasing. 

\subsubsection{Pose Estimation Results}
\begin{figure}
  \centering
   \includegraphics[width=1.0\linewidth]{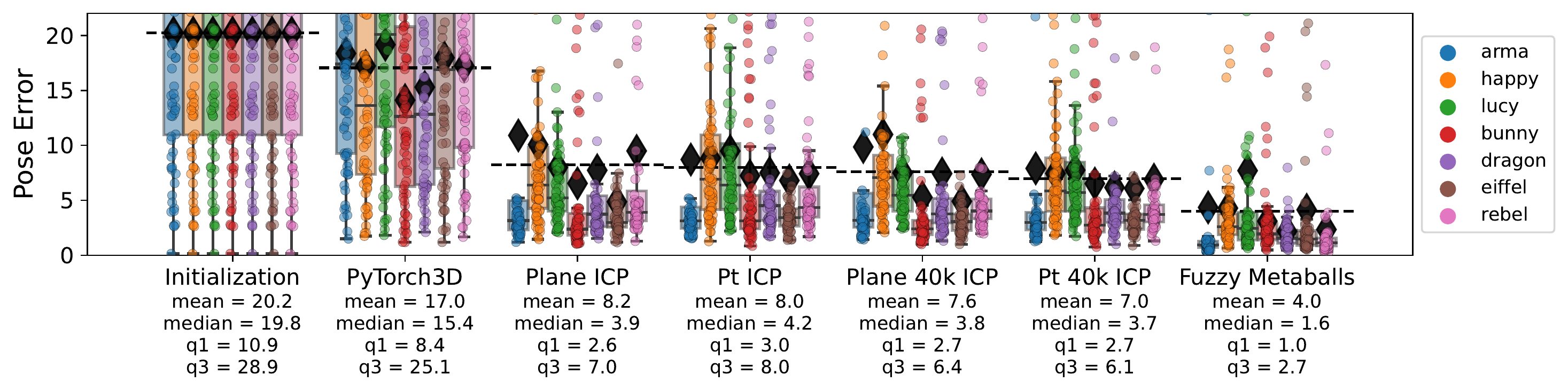}
    \caption{\textbf{Noisy Pose Estimation} Dashed lines are averages for the method, while the black diamonds show the average for that method and model. Here Fuzzy Metaballs win in all statistical measures, typically by a factor of $\approx 2$. }
    \label{fig:noisy}
\end{figure}

Overall results are found in \cref{tab:pose_est_results} and a more detailed breakdown in ~\cref{fig:noisy}. All methods sometimes struggle to find the correct local minima in this testing setup. Prior differentiable renderers significantly under-performed classic baselines like ICP, while our approximate renderer even outperforms the ICP baselines under realistic settings with synthetic noise.  

ICP on noise-free data had bimodal results: it typically either recovered the correct pose to near machine precision or it fell into the wrong local minima. Despite having a higher mean error, ICP's median errors on noise-free data were $\frac{1}{10}$ of Fuzzy Metaballs (FMs). With noisy data, this bimodal distribution disappears and Fuzzy Metaballs outperform on all tested statistical measures. FMs even outperformed ICP with high-fidelity point clouds, suggesting a difference in method not just fidelity. This may be due to our inclusion of a silhouette loss, the benefits of projective correspondence over the nearest neighbors used by this ICP variant~\cite{Zhou2018} or the strengths of visual loss over geometric loss~\cite{bundle2000triggs}.

\subsection{3D Reconstruction}\label{sec:recon}
Reconstruction experiments are common in the differential rendering literature~\cite{lassner2020pulsar,DSS_points}.  However, instead of optimizing with annotations of color~\cite{Laine2020diffrast} or normals~\cite{DSS_points}, we instead only optimize only over silhouettes, as in the classic Shape From Silhouette (SFS) ~\cite{cheung2005shape}. Unlike many prior examples in the literature, which require fine-tuning of several regularization losses~\cite{ravi2020pytorch3d,DSS_points}, we use no regularization in our experiments and can keep constant settings for all objects. 

We initialize with a sphere (isosphere for meshes, an isotropic Gaussian of points for point clouds and a small blobby sphere for Fuzzy Metaballs). Given a set of silhouette images and their camera poses, we then optimize silhouette loss for the object. In our experiments, we use 64 x 64 pixel images and have 32 views. For these experiments, we resize all models to a canonical scale and use the Adam~\cite{kingma2014Adam} optimizer. For baseline hyperparameters, we use the PyTorch3D settings with minimal modification. For SoftRas, we use a twice subdivided icosphere. For NeRF~\cite{nerf20}, we use a two layer MLP with 30 harmonic function embedding with 128 hidden dimension and the same early exit strategy as FMs. 

Inspired by artifacts seen in real videos (\cref{fig:moose_examples}), we  produce a noisy silhouette dataset where training data had $\frac{1}{8}$ of each silhouette under-segmented (\cref{fig:sill_holes}) in 16 of 32 images by clustering silhouette coordinates~\cite{minibatchKmeans2010} and removing a cluster.

\begin{figure}[tbh!]
     \centering
     \begin{subfigure}[t]{0.24\linewidth}
         \centering
         \fbox{ \centering \includegraphics[trim={0.75cm 0.75cm 0.75cm 0.75cm},clip,width=0.85\linewidth]{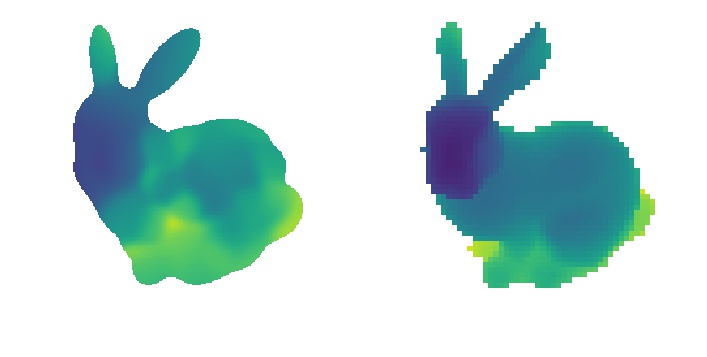} }
     \end{subfigure}
     \begin{subfigure}[t]{0.24\linewidth}
         \centering
         \fbox{ \centering \includegraphics[trim={0.75cm 0.75cm 0.75cm 0.75cm},clip,width=0.85\linewidth]{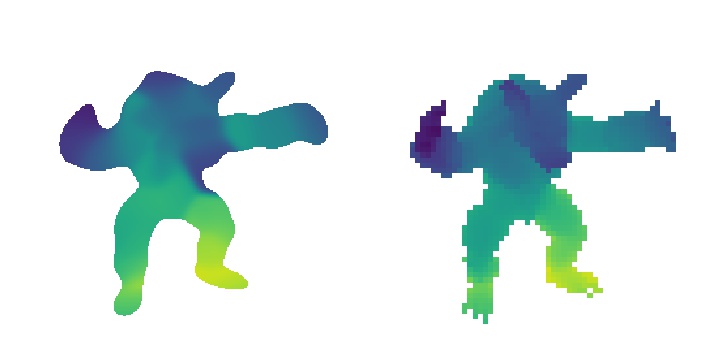} }
     \end{subfigure}
     \begin{subfigure}[t]{0.24\linewidth}
         \centering
         \fbox{ \centering \includegraphics[trim={0.75cm 0.75cm 0.75cm 0.75cm},clip,width=0.85\linewidth]{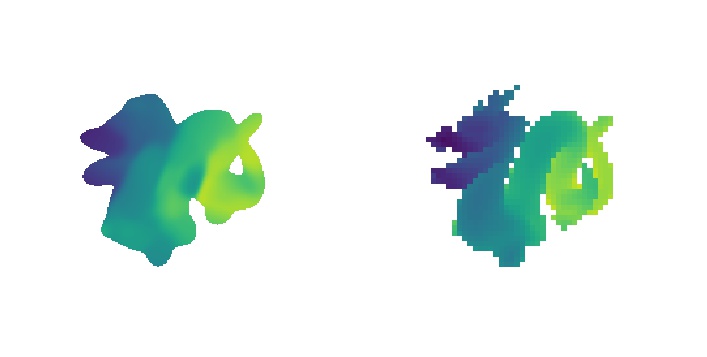} }
     \end{subfigure}
         \begin{subfigure}[t]{0.24\linewidth}
         \centering
         \fbox{ \centering \includegraphics[trim={0.75cm 0.75cm 0.75cm 0.75cm},clip,width=0.85\linewidth]{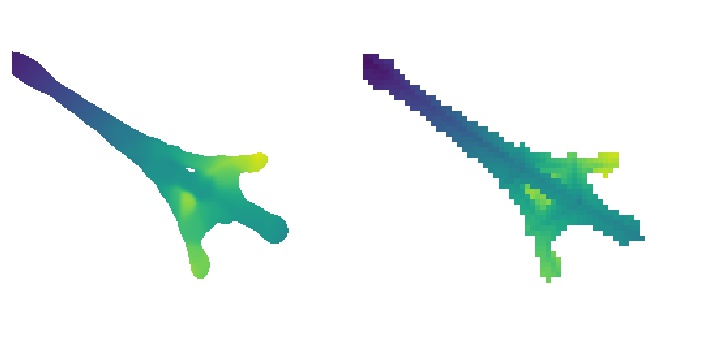} }
     \end{subfigure}
    \vfill
       \begin{subfigure}[t]{0.24\linewidth}
         \centering
         \fbox{ \centering \includegraphics[trim={0.75cm 0.75cm 0.75cm 0.75cm},clip,width=0.85\linewidth]{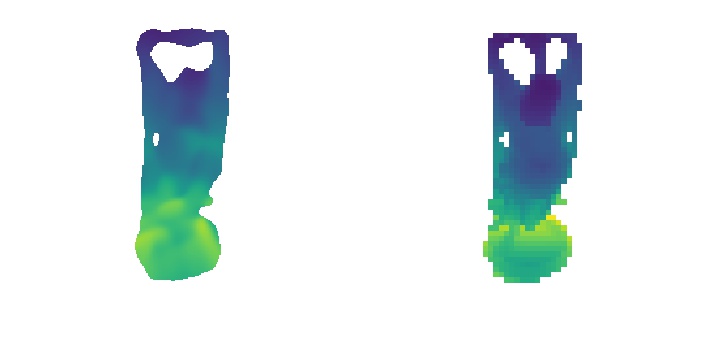} }
     \end{subfigure}
     \begin{subfigure}[t]{0.24\linewidth}
         \centering
         \fbox{ \centering \includegraphics[trim={0.75cm 0.75cm 0.75cm 0.75cm},clip,width=0.85\linewidth]{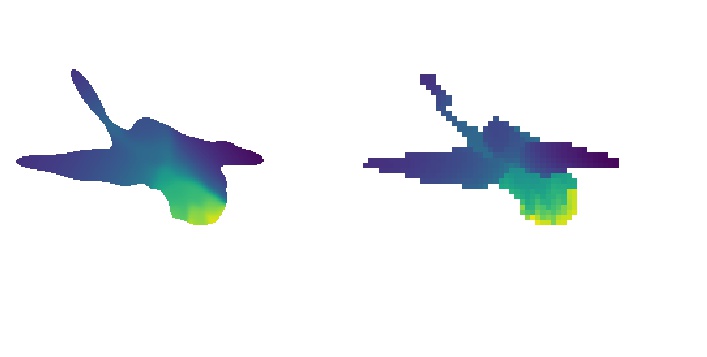} }
     \end{subfigure}
        \begin{subfigure}[t]{0.24\linewidth}
         \centering
         \fbox{ \centering \includegraphics[trim={0.75cm 0.75cm 0.75cm 0.75cm},clip,width=0.85\linewidth]{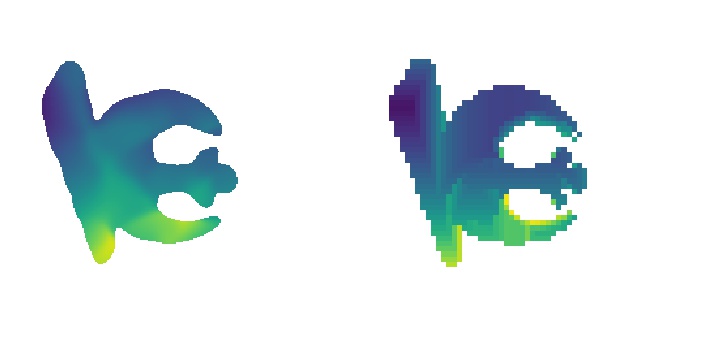} }
     \end{subfigure}
                 \begin{subfigure}[t]{0.24\linewidth}
         \centering
         \fbox{ \centering \includegraphics[trim={0.75cm 0.75cm 0.75cm 0.75cm},clip,width=0.85\linewidth]{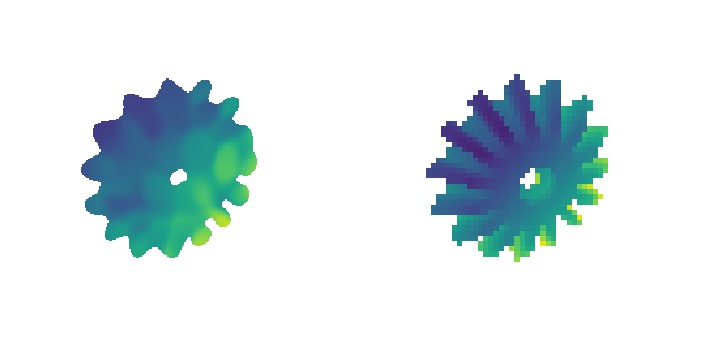} }
     \end{subfigure}

    \caption{\textbf{Shape from Silhouette (SFS)} reconstructions. On the left is a 40 component Fuzzy Metaball result and the right is the mesh ground-truth of about 2,500 faces, both colored by depth maps. }
    \label{fig:sill_results}
\end{figure}

\begin{table}[tb]
\centering
\caption{\textbf{Shape from Silhouette reconstruction fidelity} as measured by cross-entropy silhouette loss on 32 novel  viewpoints for each of 10 sample models.  Runtimes were the average per model and performed on CPU. Results show $\mu \pm \sigma$. }
\begin{tabular}{@{}p{0.27\textwidth}C{0.12\textwidth}C{0.30\textwidth}C{0.25\textwidth}@{}}
\toprule
 & \textbf{Time (s)} & \textbf{ Noise-Free Recon. Error} & \textbf{Noisy Recon. Error} \\ \midrule
Voxel Carving ~\cite{martin1983silhouette,Zhou2018} & \underline{82} & 0.31 $\pm$ 0.100 & $1.119 \pm 0.367$ \\
PyTorch3D Point ~\cite{ravi2020pytorch3d} & 185 & 0.075 $\pm$ 0.066 & $0.100 \pm 0.079$ \\
PyTorch3D Mesh ~\cite{liu2019soft} & 3008 & 0.062 $\pm$ 0.049 & $0.072 \pm 0.051$\\
NeRF ~\cite{nerf20} & 7406 & \textbf{0.032} $\pm$ \textbf{0.022} & \underline{0.062 $\pm$ 0.063} \\ \midrule
Fuzzy Metaballs & \textbf{68} & \underline{0.040 $\pm$ 0.015} & \textbf{0.055} $\pm$ \textbf{0.016} \\ 
\bottomrule
\end{tabular}

\label{tab:quant_recon}
\end{table}

\subsubsection{Shape From Silhouette Results} We show qualitative reconstructions from Fuzzy Metaballs (\cref{fig:sill_results}), along with quantitative results against baselines (\cref{tab:quant_recon}) and some example reconstructions from all methods (\cref{fig:sill_holes}). 

Overall, we found that our method was significantly faster than the other differentiable renderers, while producing the best results in the case of noisy reconstructions. Classic voxel carving~\cite{martin1983silhouette} with a $384^3$ volume was reasonably fast, but the 32 views of low resolution images didn't produce extremely sharp contours  (\cref{fig:sill_holes_noisefree}). With under-segmentation noise, voxel carving fails completely while the differentiable renderers reasonably reconstruct all models. 

Among the differentiable renders, we can see how the mesh-based approach struggles to change genus from a sphere to the Eiffel tower. The point cloud renderer lacks the correct gradients to successfully pull spurious points into the model. NeRF~\cite{nerf20} performs reasonably well in shape from silhouette, even with spurious masks. In fact, it was the best performer for noise-free data, and in a majority of the reconstructions in noisy data (its mean performance was hurt by results on \texttt{eiffel} and \texttt{lucy} with long thin surfaces). NeRF is a sophisticated model with many settings, and it may have a configuration where it successfully reconstructs all the models, but due to its dense volumetric rendering and use of an MLP, it is 100x slower than our low degree of freedom representation. 

\begin{figure}[htb!]
          \begin{subfigure}[t]{0.10\linewidth}
         \centering
         \includegraphics[width=1.0\linewidth]{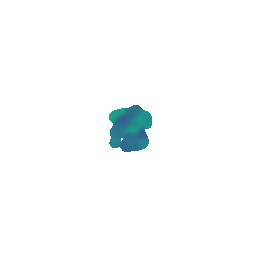}
     \end{subfigure}
          \begin{subfigure}[t]{0.10\linewidth}
         \centering
         \includegraphics[width=1.0\linewidth]{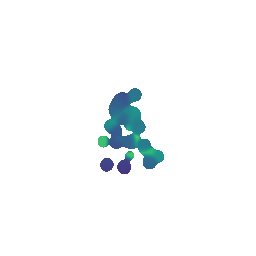}
     \end{subfigure}
    \begin{subfigure}[t]{0.10\linewidth}
         \centering
         \includegraphics[width=1.0\linewidth]{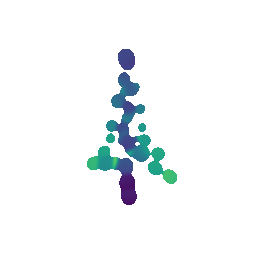}
     \end{subfigure}
               \begin{subfigure}[t]{0.10\linewidth}
         \centering
         \includegraphics[width=1.0\linewidth]{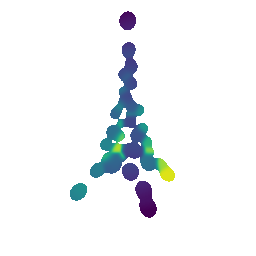}
     \end{subfigure}
     \begin{subfigure}[t]{0.10\linewidth}
         \centering
         \includegraphics[width=1.0\linewidth]{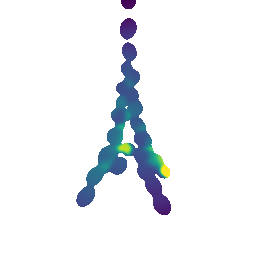}
     \end{subfigure}
     \begin{subfigure}[t]{0.10\linewidth}
         \centering
         \includegraphics[width=1.0\linewidth]{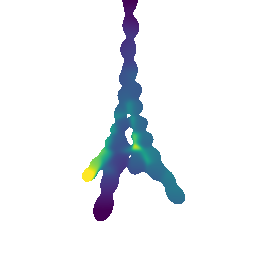}
     \end{subfigure}
     \begin{subfigure}[t]{0.10\linewidth}
         \centering
         \includegraphics[width=1.0\linewidth]{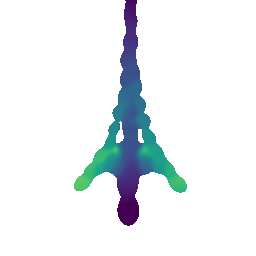}
     \end{subfigure}
      \begin{subfigure}[t]{0.10\linewidth}
         \centering
         \includegraphics[width=1.0\linewidth]{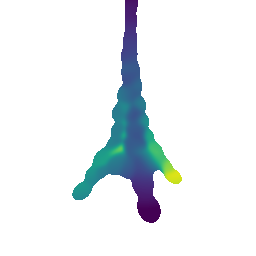}
     \end{subfigure}
     \begin{subfigure}[t]{0.10\linewidth}
         \centering
         \includegraphics[width=1.0\linewidth]{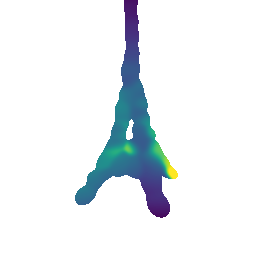}
     \end{subfigure}

\vfill

          \begin{subfigure}[t]{0.10\linewidth}
         \centering
         \includegraphics[width=1.0\linewidth]{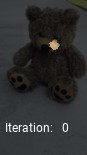}
     \end{subfigure}
          \begin{subfigure}[t]{0.10\linewidth}
         \centering
         \includegraphics[width=1.0\linewidth]{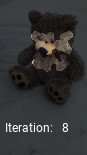}
     \end{subfigure}
    \begin{subfigure}[t]{0.10\linewidth}
         \centering
         \includegraphics[width=1.0\linewidth]{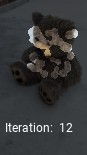}
     \end{subfigure}
               \begin{subfigure}[t]{0.10\linewidth}
         \centering
         \includegraphics[width=1.0\linewidth]{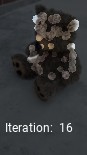}
     \end{subfigure}
     \begin{subfigure}[t]{0.10\linewidth}
         \centering
         \includegraphics[width=1.0\linewidth]{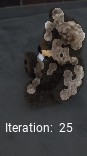}
     \end{subfigure}
     \begin{subfigure}[t]{0.10\linewidth}
         \centering
         \includegraphics[width=1.0\linewidth]{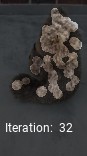}
     \end{subfigure}
     \begin{subfigure}[t]{0.10\linewidth}
         \centering
         \includegraphics[width=1.0\linewidth]{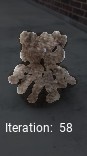}
     \end{subfigure}
      \begin{subfigure}[t]{0.10\linewidth}
         \centering
         \includegraphics[width=1.0\linewidth]{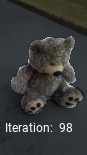}
     \end{subfigure}
     \begin{subfigure}[t]{0.10\linewidth}
         \centering
         \includegraphics[width=1.0\linewidth]{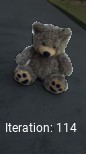}
     \end{subfigure}

    \caption{\textbf{Shape from Silhouette steps} Top row shows synthetic data with reconstructed depth. Bottom row shows reconstructed masks for a CO3D video~\cite{reizenstein21co3d}. }
    \label{fig:co3d}
\end{figure}

\begin{figure}[tbh!]
     \centering
     \begin{subfigure}[t]{0.93\linewidth}
         \centering
         \includegraphics[width=1.0\linewidth]{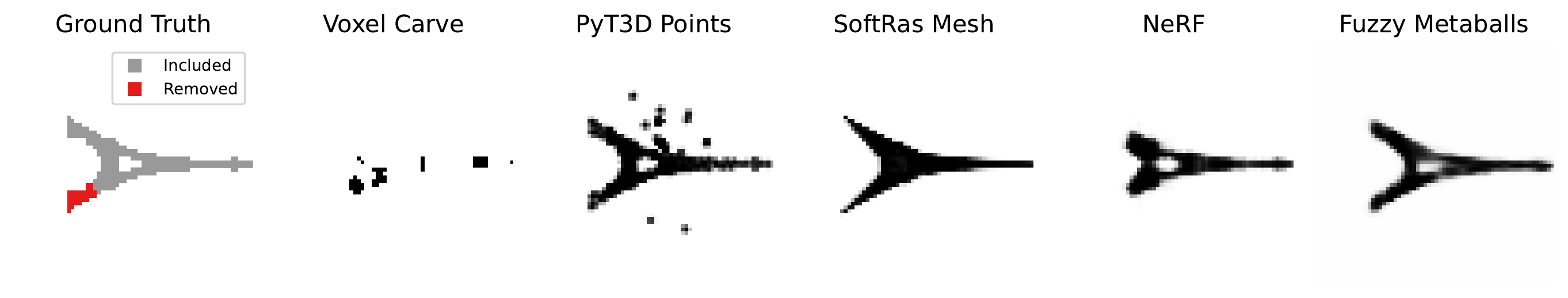}
     \end{subfigure}
     \vfill
     \begin{subfigure}[t]{0.93\linewidth}
         \centering
         \includegraphics[width=1.0\linewidth]{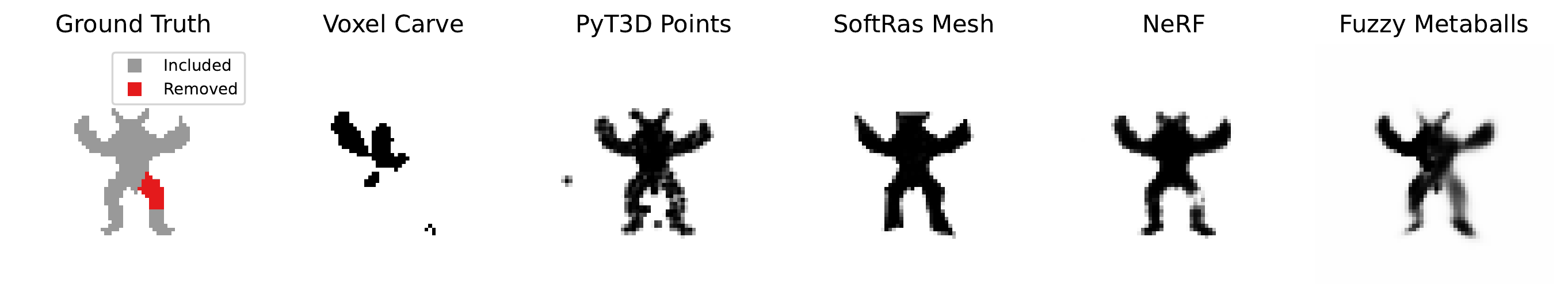}
     \end{subfigure}
    \caption{\textbf{Shape from Silhouette Results} with simulated under-segmentation. }
    \label{fig:sill_holes}
\end{figure}

\begin{figure}[t]
     \centering
          \begin{subfigure}[t]{1.0\linewidth}
         \centering
   \includegraphics[width=0.92\linewidth]{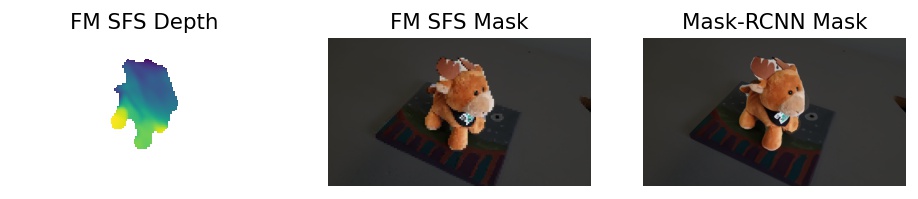}
         \caption{\textbf{Depth and silhouette} from a shape-from-silhouette reconstruction. }
     \end{subfigure}
          \begin{subfigure}[t]{1.0\linewidth}
         \centering
         \includegraphics[width=0.92\linewidth]{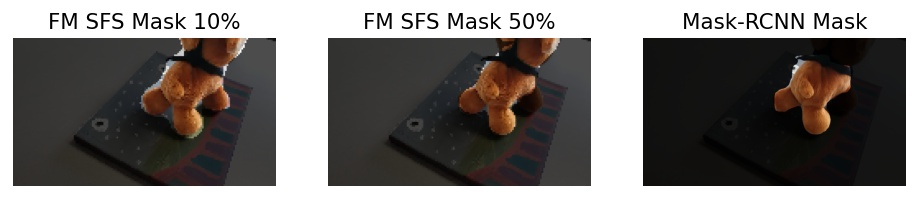}
         \caption{\textbf{Recovering from undersegmentation in the ground truth masks.} While a 50\% threshold does a good job recovering the head, better recovery can be shown with a 10\% threshold, also recovering the leg. }
     \end{subfigure}
          \begin{subfigure}[t]{1.0\linewidth}
         \centering
         \includegraphics[width=0.92\linewidth]{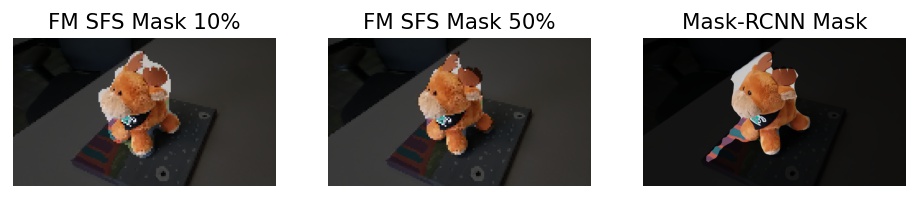}
         \caption{\textbf{Recovering from oversegmentation in ground truth masks.} Even the $\alpha=10\%$ threshold only leads to minor over-segmentation in the mask, suggesting a setting that be appropriate in general. }
     \end{subfigure}

    \caption{\textbf{Shape from silhouette reconstruction on natural images} from a handheld cell phone video, using COLMAP~\cite{schoenberger2016mvs} and Mask RCNN~\cite{He2017MaskR} for automatic camera poses and silhouettes. The low degree of freedom leads to natural regularization and recovery from  errors in ground truth.   }    \label{fig:moose_examples}
\end{figure}

\section{Discussion}
The focus of our approximate differentiable rendering method has been on shape. While it is possible to add per-component colors to Fuzzy Metaballs (\cref{fig:color_blend}), that has not been the focus of our experiments. Focusing on shape allows us to circumvent modeling high-frequency color textures, as well as ignoring lighting computations. This shape-based approach can use data from modern segmentation methods~\cite{He2017MaskR} and depth sensors~\cite{DBLP:journals/corr/KeselmanWGB17}. Low-degree of freedom models have a natural robustness and implicit regularization that allows for recovery from significant artifacts present in real systems. For example, \cref{fig:moose_examples} shows robust recovery from real over/under-segmentation artifacts in video sequences. 

Our approximate approach to rendering by using OIT-like methods creates a trade-off. The downside is that small artifacts can be observed since the method coarsely approximates correct image formation. The benefits are good gradients, speed \& robustness, all of which produce utility in vision tasks. 

Compared to prior work~\cite{lassner2020pulsar,liu2019soft}, our results do not focus on the same areas of differentiable rendering. Unlike other work, we do not perform GPU-centric optimizations~\cite{Laine2020diffrast}. Additionally, prior work focuses on producing high-fidelity color images (and using them for optimization). Unlike prior work, we benchmark our method across a family of objects and report quantitative results against classic baselines. Unlike some popular implicit surface methods such as the NeRF~\cite{nerf20} family, our object representation is low degree of freedom, quick to optimize from scratch, and all the parameters are interpretable with geometric meaning. 

While our experiments focus on classic computer vision tasks such as pose estimation or shape from silhouette, the value of efficiently rendering interpretable, low degree of freedom models may have the biggest impact outside of classic computer vision contexts. For example, in scientific imaging it is often impossible to obtain high-quality observations since the sensors are limited. For example, in  non-light-of-sight imaging~\cite{Tsai_2019_CVPR}, sonar reconstruction~\cite{9197042}, lightcurve inversion~\cite{kaasalainen2001optimization} and CryoEM~\cite{cryoEM2015,Zhong_2021_ICCV}. In all these contexts, getting good imaging information is extremely hard and low degree of freedom models could be desirable. 

\section{Conclusion}
Approximate differentiable rendering with algebraic surfaces enables fast analysis-by-synthesis pipelines for vision tasks which focus on shapes, such as pose estimation and shape from silhouette. For both tasks, we show results with realistic, simulated noise. The robustness of our approach enables it to runs naturally on silhouettes extracted from real video sequences without any regularization. Whereas classic methods can struggle once noise is introduced, differentiable renderers naturally recovery by using stochastic optimization techniques. By using gradient-based optimization, differentiable rendering techniques provide a robust solution to classic vision problems.  Fuzzy Metaballs can enable low-latency differential rendering on CPUs. Our formulation connects algebraic surfaces~\cite{10.1145/357306.357310} used in graphics with Gaussian Mixture Models~\cite{Eckart2016} used in vision. These provide a compact, interpretable representation for shape with many uses. 

%
%
\bibliographystyle{splncs04}
\bibliography{egbib}

\clearpage
\appendix

\begin{figure}[t]
     \centering
     \begin{subfigure}[t]{0.48\linewidth}
         \centering
         \includegraphics[width=1.0\linewidth]{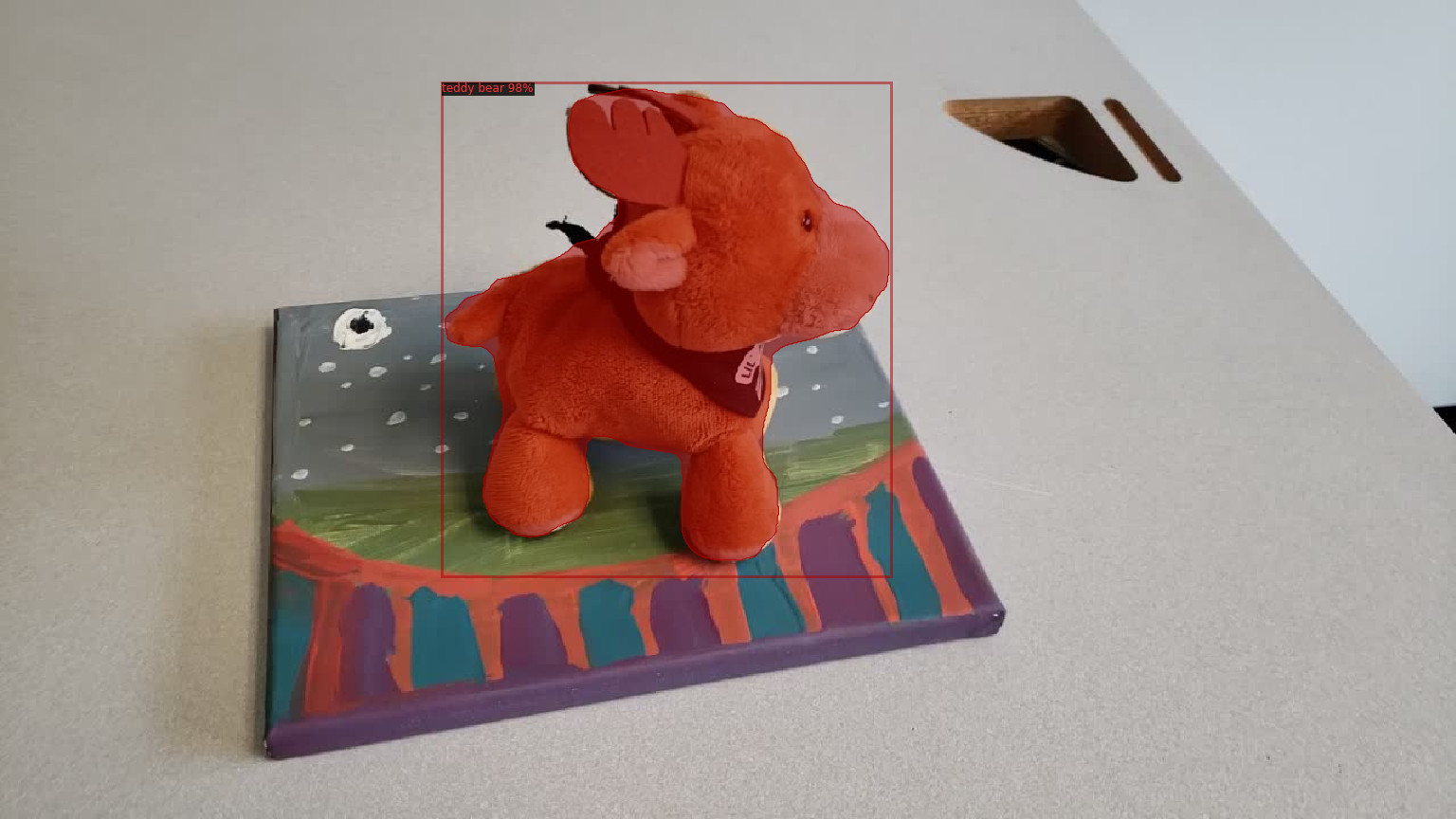}
         \caption{Mask RCNN output for valid frame}
     \end{subfigure}
     \hfill
     \begin{subfigure}[t]{0.48\linewidth}
         \centering
         \includegraphics[width=0.7\linewidth]{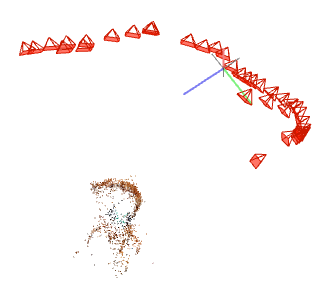}
         \caption{COLMAP estimate of camera poses}
     \end{subfigure}
    \begin{subfigure}[t]{0.48\linewidth}
         \centering
         \includegraphics[width=1.0\linewidth]{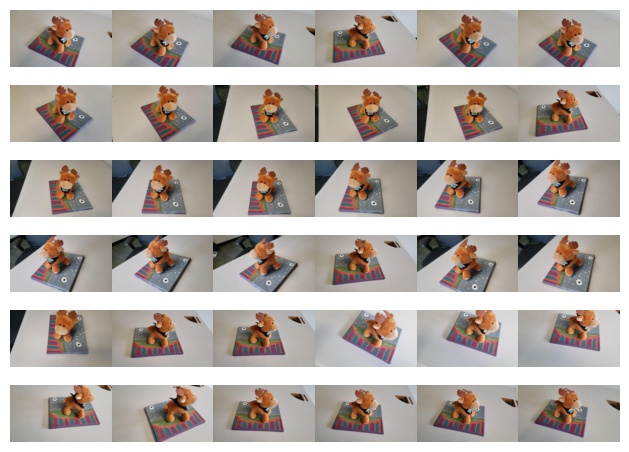}
         \caption{All 36 frames used for SFS}
     \end{subfigure}
    \begin{subfigure}[t]{0.48\linewidth}
         \centering
         \includegraphics[width=1.0\linewidth]{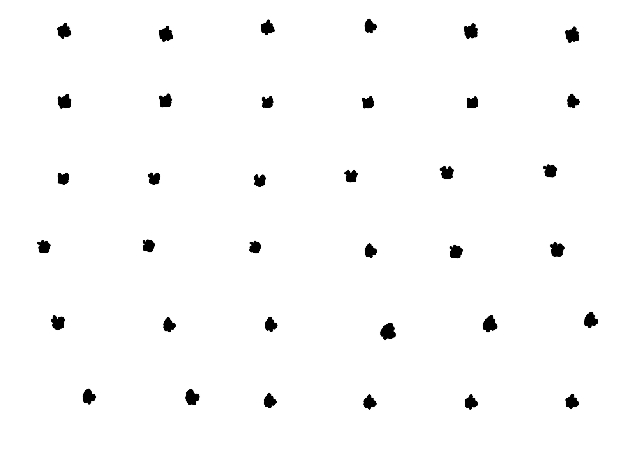}
         \caption{SFS initialization}
     \end{subfigure}
         \begin{subfigure}[t]{0.48\linewidth}
         \centering
         \includegraphics[width=1.0\linewidth]{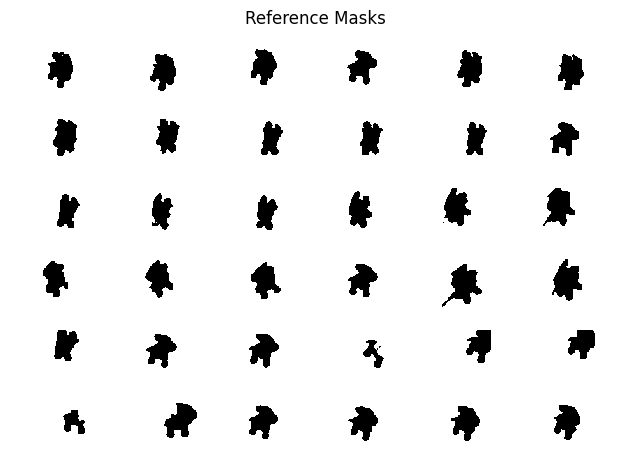}
         \caption{Mask RCNN Silhouettes}
     \end{subfigure}
         \begin{subfigure}[t]{0.48\linewidth}
         \centering
         \includegraphics[width=1.0\linewidth]{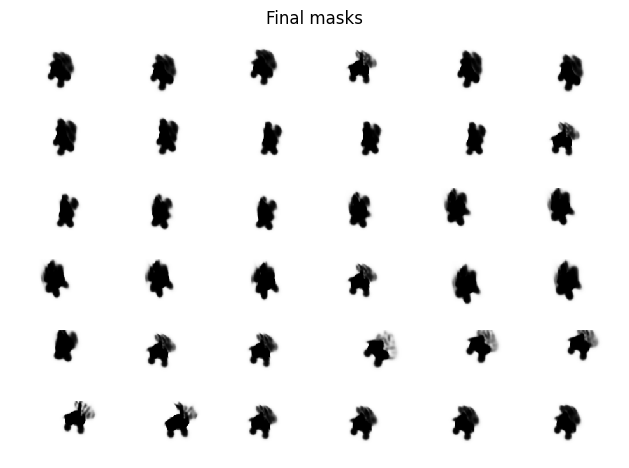}
         \caption{SFS Mask Results}
     \end{subfigure}
    \caption{\textbf{Video-based SFS reconstruction}}
    \label{fig:moose_shown}
\end{figure}

\section{Video Results}
Here we describe additional details about the experiment shown in Fig. 9 of the main paper. Concerning the differentiable renderer: the method, settings and hyperparameters are identical to those in Fig. 7 and 8 and Section 5.3. We simply run the method on different input. 

We collected a 14 second video with a Samsung S9 cell phone at 1280 x 720 resolution at 30 Hz. The video contains motion blur, auto-exposure, and clearly visible video compression artifacts, making it unsuitable for some reconstruction methods. We sub-sampled the video down to 6Hz and ran Mask RCNN~\cite{He2017MaskR} from Detectron2~\cite{wu2019detectron2} with the pre-trained weights \texttt{COCO-InstanceSegmentation} \texttt{/mask\_rcnn\_R\_50\_FPN\_3x.yaml} to detect objects. In our case, the object was detected as part of the teddy bear class, with about 55 viable frames. We ran COLMAP~\cite{schoenberger2016mvs} to obtain camera poses for those frames, where COLMAP successfully returned 36 frames with valid camera poses. We ran our SFS pipeline at 160 x 90 resolution to obtain the results shown. Visual examples from this pipeline are shown in \cref{fig:moose_shown}. All the methods used their default settings; there was no parameter tuning involved. 

\subsection{Video Result Analysis}
The trajectory shown here only covers about half of the object from a roughly constant elevation. Complicating the reconstruction is that the camera poses are imperfect due to estimation and unmodeled camera distortion. Much more significant is that the Mask RCNN silhouettes used for reconstruction are often extremely under or over segmented. 

Despite these issues in the "ground truth" used for optimization, the low degree of freedom of Fuzzy Metaballs allows the model to reasonably recover from the massive artifacts. While the result in the main paper shows the default 50\% threshold, to recover some areas, we have to lower our $\alpha$ threshold to 10\%. 
\clearpage
\section{Hyper-parameters}
Our proposed method has 5 hyper-parameters described in the paper. Briefly, $\beta_1$ prioritizes close hits, $\beta_2$ prioritizes hits closer to the camera,   $\beta_3$ prioritizes hits in front of the camera, and $\beta_4$ and $\beta_5$ serve as a scale and offset to generate alpha masks. 
Since our system is fully algebraic, it is possible to perform gradient descent into these hyper-parameters (and the functional form of JAX naturally returns their gradients), but this was not done.

Instead, we optimized them for depth and alpha mask accuracy over a small simulated dataset of the Stanford bunny using standard black-box optimization techniques~\cite{hansen2016cma,hansen2019pycma,loshchilov2016cmaes} before running most of our experiments. We found that the ray-based renderer led to similar optimal hyperparameters across multiple tested resolutions, across a wide range of mixture components, and across our linear, quadratic and cubic methods of intersection computation. 
\begin{figure}[thb]
  \centering
   \includegraphics[width=1.0\linewidth]{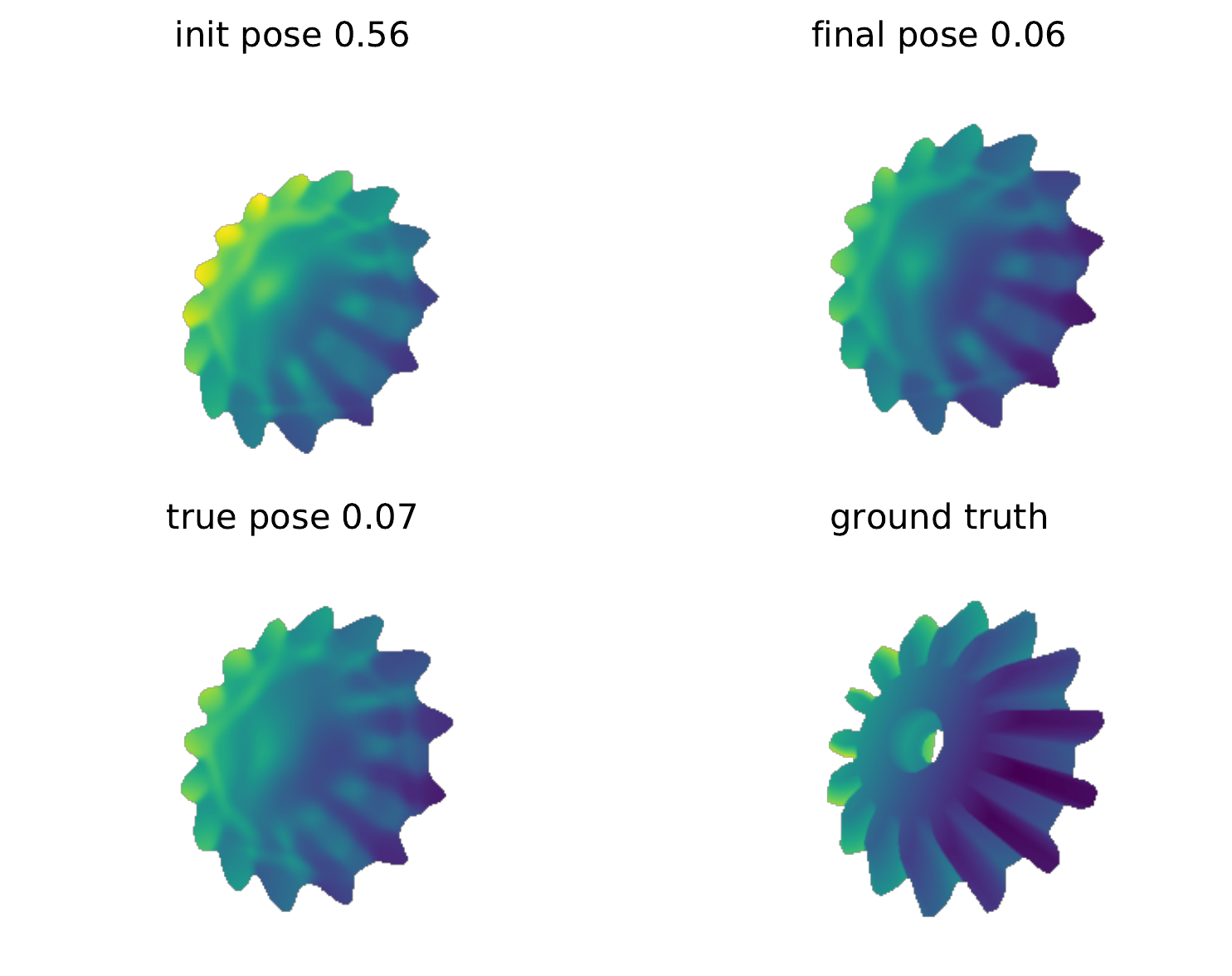}
    \caption{\textbf{Gear Results with Fuzzy Metaballs} Final pose describes the final pose after gradient-based pose optimization, while true pose is rendered view from the ground truth pose. Ground truth is the Blender-generated depth map of the full-fidelity model. The final pose shown here has a rotational error of 23.9 degrees. However, the gear has 15 teeth and hence a 24.0 degree symmetry. }
    \label{fig:gear}
\end{figure}
\section{Exclusion of \texttt{gear} model}
The \texttt{gear} model was selected because of its interesting geometry from Thingi10k~\cite{DBLP:journals/corr/ZhouJ16}. However, for pose estimation, we exclude its results from the overall average due to symmetry. Our poses are generated with rotations of uniform axis and angle uniformly between -45 and 45 degrees (uniform-axis random spin~\cite{random_rotation_thesis}). The gear model however has 15 teeth and a rotational symmetry of 24 degrees when viewed from one side, as seen in \cref{fig:gear}. This can sometimes produce pose errors with no real geometric error. 

The model itself is not symmetric, with 15 gears and a back face with 180 degree symmetry. But with a single view, our testing conditions can generate poses which are geometrically correct but produce pose errors. The other model with symmetry, \texttt{eiffel}, only has 90 degree symmetry and our testing conditions place all random poses in the same local minima.  

We don't use the \texttt{yoga} or \texttt{plane} models for pose estimation as we only latter added them for the reconstruction experiments. Both models originate from prior differentiable rendering uses in reconstruction ~\cite{liu2019soft,DSS_points}.

\section{Pose Estimation Details}\label{sec:pose_est_supp}
\begin{figure*}[th]
  \centering
   \includegraphics[width=1.0\linewidth]{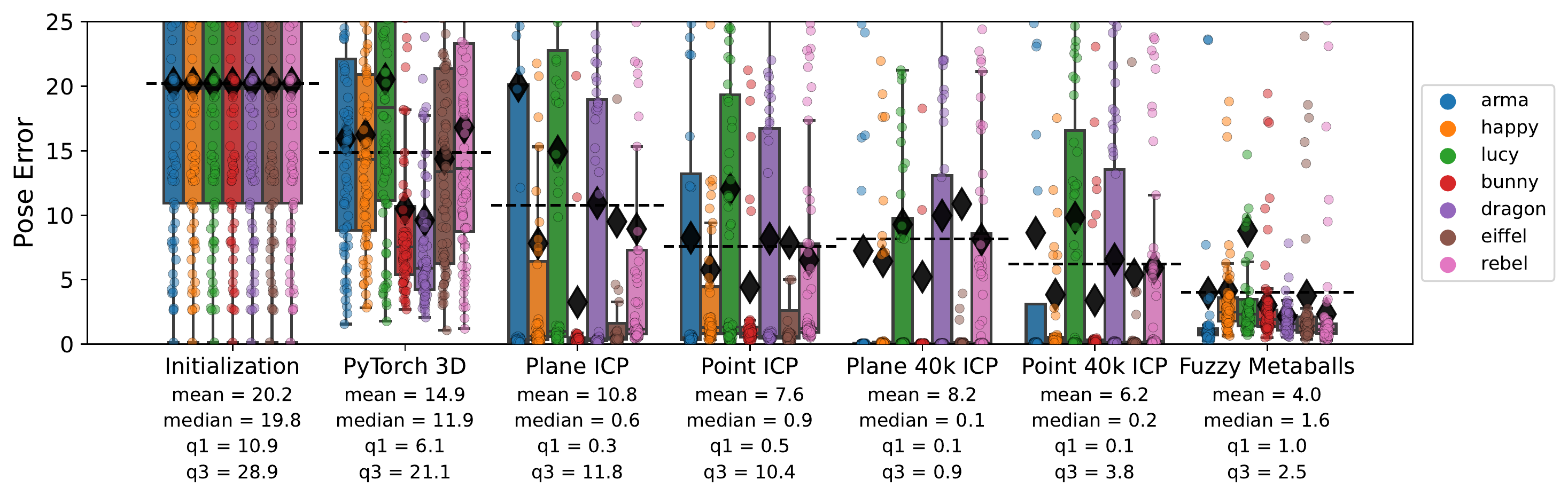}
   \includegraphics[width=1.0\linewidth]{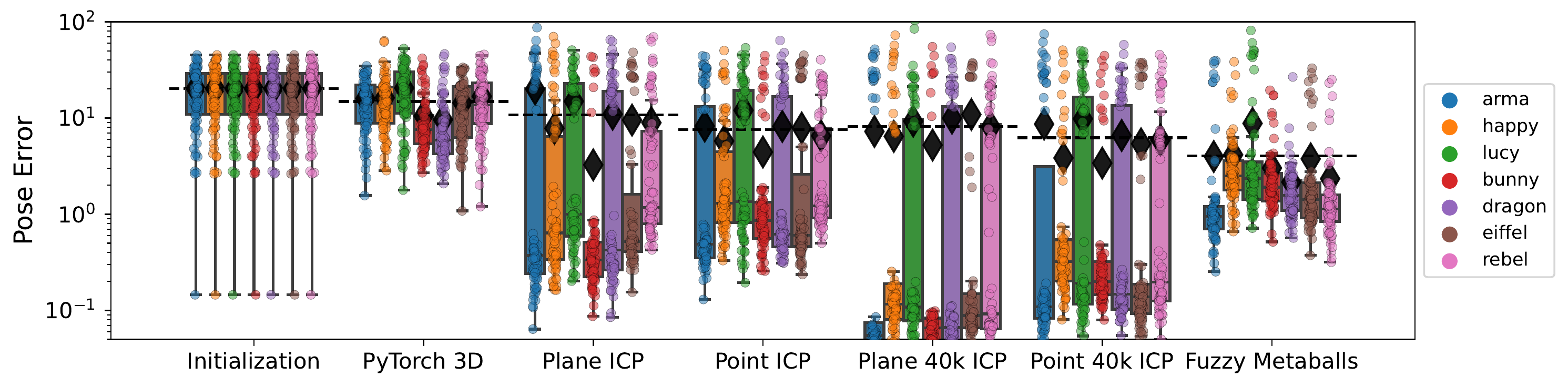}
    \caption{\textbf{Noise Free Pose Estimation} Linear scale plot above and log-scale below. Dashed lines are averages for the method, while the black diamonds show the average for that method and model. Statistics for each method are listed. \texttt{gear} model is excluded from statistics.  }
    \label{fig:noise_free}
\end{figure*}

We include noise-free results the same seed as the noisy results in the main paper. Summary plots are shown in \cref{fig:noise_free} and \cref{fig:noisy}. In the noise-free case, we find that Point-to-Point ICP works better. With noise, Point-to-Plane ICP methods perform better.

\subsection{Noise Free}\label{sec:noise-free}
As described in the paper, when ICP methods perform well, they perform extremely well, an order of magnitude better than the differentiable renderers (see the log-scale plot), to fractions of a degree since they have high resolution samples. However, sometimes ICP finds poor local minima and on average our method performs better, even when ICP has a dense point cloud.  Despite having a better mean, Fuzzy Metaballs (FM) have a median error that is 8 times higher and a 25th percentile error that is 10 times higher. The increase in robustness from FM is demonstrated in lower 75th percentile errors. 

\subsection{Noisy Depth Images}\label{sec:noisy}
With synthetic noise, both differentiable renderer methods are barely affected, while the ICP results see a large degradation in peak performance. Under this experimental condition, Fuzzy Metaballs have the lowest mean, median, 25th and 75th percentile errors (typically by a factor of 2 compared to ICP).

Interestingly, some of the worst case performance of the ICP methods disappears (lower q3 measurements) when noise is added. We hypothesize that this occurs due to a form of symmetry breaking that helps avoid singularities and bad correspondences. Fuzzy Metaballs, being a low fidelity model, experience nearly no degradation in performance when noise is added to depth images. 

\begin{figure*}[th]
  \centering
   \includegraphics[width=1.0\linewidth]{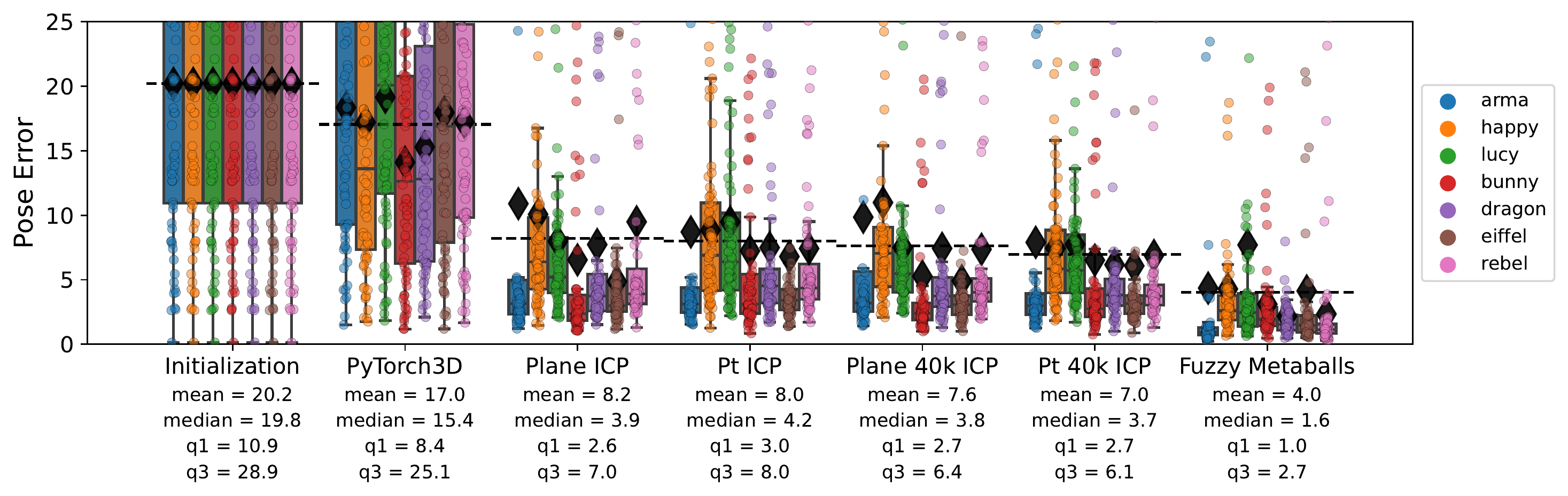}
   \includegraphics[width=1.0\linewidth]{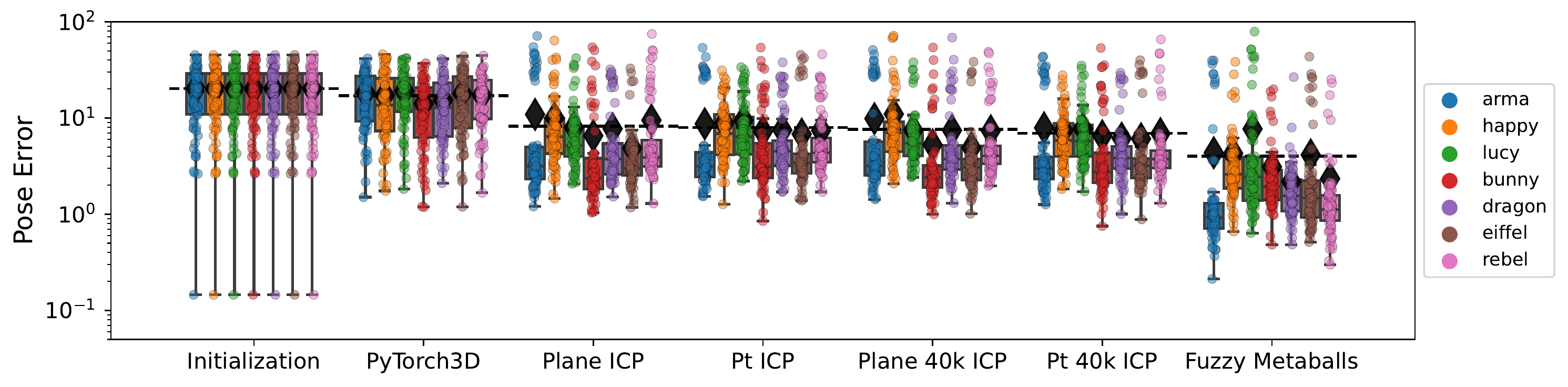}
    \caption{\textbf{Noisy Pose Estimation} Identical visualization to \cref{fig:noise_free}. Linear scale figure is identical to that in the main paper.  }
    \label{fig:noisy_supp}
\end{figure*}
\clearpage

\begin{figure}[t]
  \centering
   \includegraphics[width=1.0\linewidth]{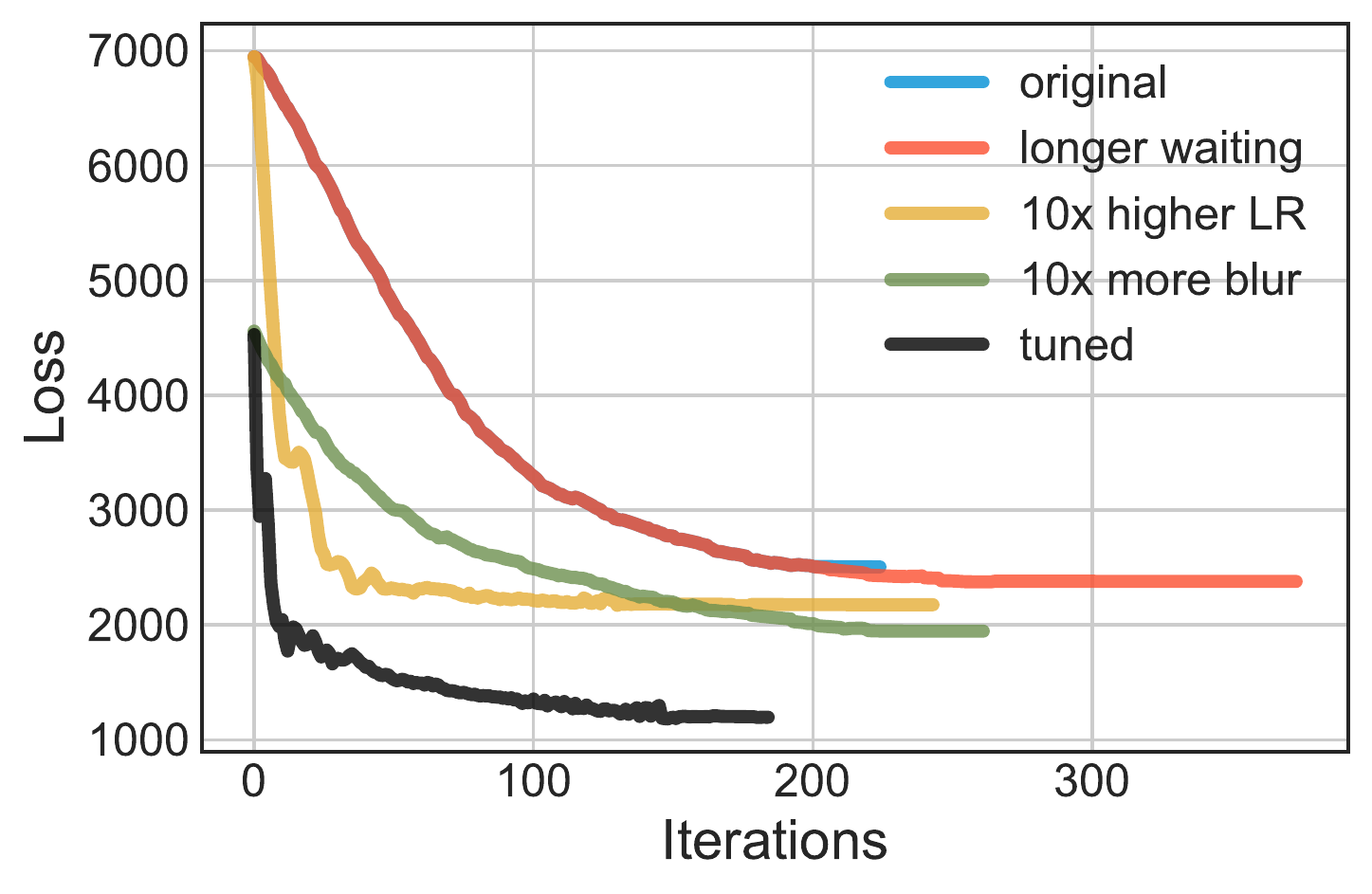}
    \caption{\textbf{PyTorch3D baseline} convergence curves using different hyperparameters for pose estimation. All curves produce roughly equivalent pose errors, significantly worse than FM or ICP. }
    \label{fig:softras_curve}
\end{figure}

\section{SoftRasterizer performance}\label{sec:tune_softras}
One might wonder about why our baseline of PyTorch3D, implementing SoftRas~\cite{liu2019soft}, performs so poorly in the pose estimation experiments. Prior work on differentiable rendering~\cite{liu2019soft,lassner2020pulsar,DSS_points} demonstrates their pose optimization experiments with mostly single, visual examples. These often use color images, and provide no baselines results from standard methods. Quantitative results in SoftRas~\cite{liu2019soft} examined solving for rotation uncertainty in images of a colored cube and their resulting rotation errors averaged over 60 degrees. It is perhaps not surprising that our pose estimation experiments, featuring a family of models, simultaneous rotation and translation, while optimizing only depth and silhouettes, might challenging these methods. 

\begin{figure}[t]
  \centering
   \includegraphics[width=0.8\linewidth]{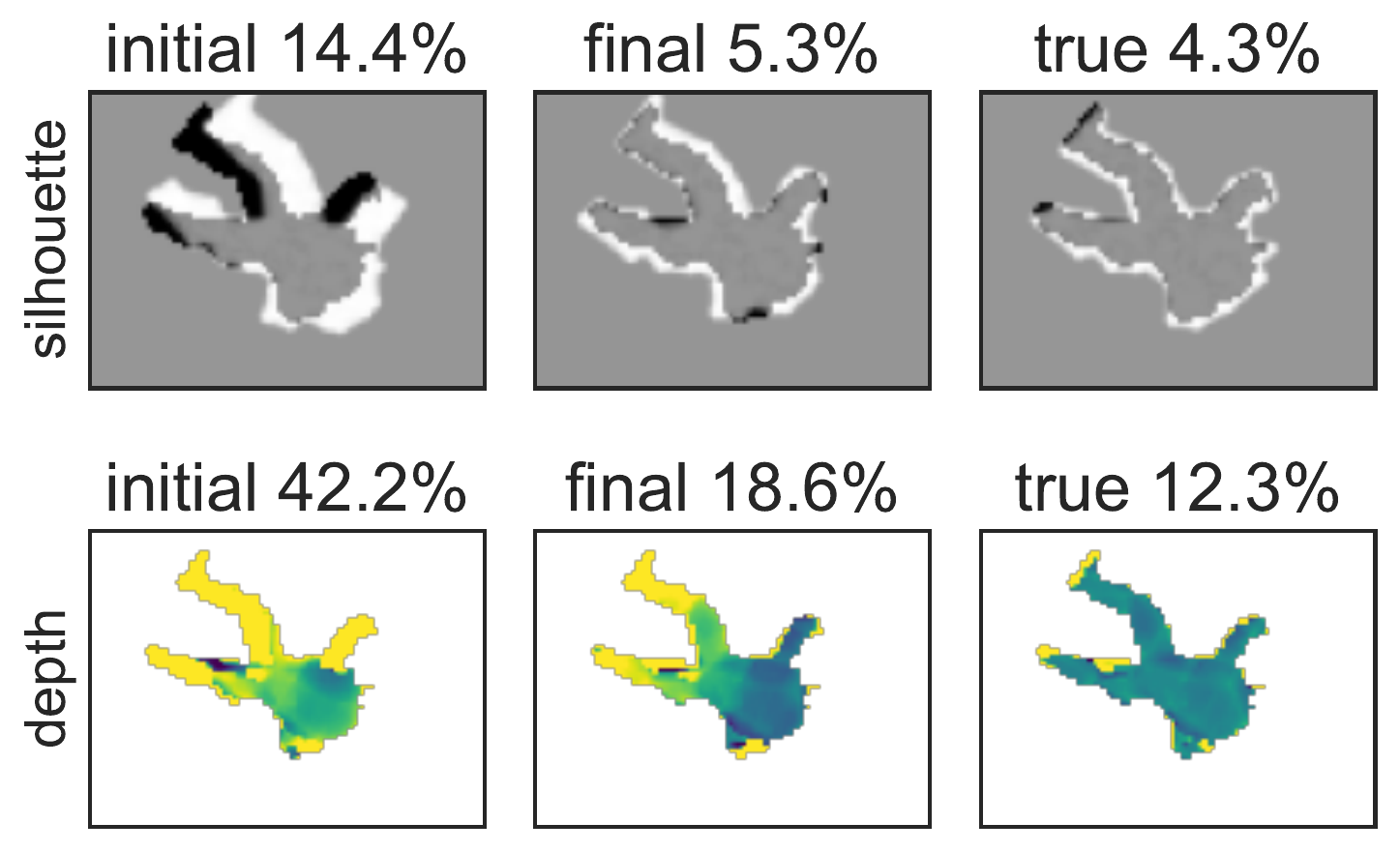}
    \caption{\textbf{PyTorch3D baseline} Visualization of errors seen with pose optimization. Initial is an example pose perturbation of the \texttt{arma} model. Final is the result after pose optimization, and true is the result of the ground truth pose. Silhouette error is in percent of pixels that are wrong, while depth error is in average relative depth error.  Optimization leads to a reasonable decrease in both.  }
    \label{fig:softras_error}
\end{figure}

We used the hyperparameters from the current PyTorch 3D~\cite{ravi2020pytorch3d} \textit{Camera position optimization} sample. We tuned learning rates to behave well with our depth + silhouette loss function, and followed an automatic learning rate schedule~\cite{king_1970}. As can be seen in \cref{fig:softras_error}, the pose optimization performs reasonably well in reducing image errors. However, the optimized pose still demonstrates visual errors compared to the ground truth pose. Even worse, the optimization perturbs the pose in such a way that the pose error at the end of optimization (16 degrees and 15\%) is worse than the pose errors at the perturbed initialization (12 degrees and 8\%), despite the significant reduction in loss. 

To check if the hyper-parameters from the PyTorch3D sample was a poor fit, we searched for settings which produced good pose estimation for a single frame. We used CMA-ES~\cite{hansen2016cma,hansen2019pycma}, a fairly common black box method~\cite{loshchilov2016cmaes}. This type of task-specific hyper-parameter optimization was \textbf{never performed} for our Fuzzy Metaballs experiments. We only performed these experiments on an existing baseline to examine how good it might perform in the best case. Convergence curves can be seen in \cref{fig:softras_curve}. 

All our manual tweaking of PyTorch3D hyper-parameters produced comparable configurations (17-18 degrees of rotation, 15-16 percent translation). The automated optimization found a setting which produced 16 degrees of rotation error and 9 percent of translation error, still worse than the perturbed initialization. However, these settings used a very high learning rate that proved unstable with other frames. Lowering to learning rate resulted in settings with a negligible improvement (2\%) to our initial settings. These tests suggest our initial hyper-parameter choices were a reasonably good setting for the baseline method. 

Further parameter search with a constraint on learning rate failed to find parameters significantly improved from the defaults. Optimization was over 8 parameters: $\sigma,\gamma$, blurring radius for both depth and silhouette, faces per pixel, learning rate, and multipliers for depth and silhouette loss. 

The results of the PyTorch3D pose optimization, when fed into the Fuzzy Metaballs renderer suggest these implementations are coherent-- the FM optimization run on the same original pose perturbation and model produces a 0.5 degree error and 1.5\% pose translation error, while even noisy point-to-point ICP gets a 2.6 degree and 4.3\% translation error. 

Both the Fuzzy Metaballs and PyTorch3D optimization use an axis-angle, 3 parameter rotation estimation. These is some evidence suggesting PyTorch Autograd for $SO(3)$ might be unstable at times in its native form~\cite{teed2021tangent}. Lastly, we suspect that the same scale invariance issues we address in \cref{sec:pose} may exist for the PyTorch3D baselines. SFS experiments were performed on objects of roughly unit scale to mitigate this potential issue. 

\subsection{Pulsar performance}
Our attempts to test a recent differentiable renderer, \textit{Pulsar}, found it performed very poorly. Not only are there software bugs with the latest PyTorch3D at the time of writing (\texttt{0.6.1}) where the code clobbers camera data-structures and requires re-creating them with every call to the render function, but the pose estimation results were very poor. 

We used the same settings as the Point Cloud Differentiable Renderer baseline we tested, which provided fair results and produced visually similar outputs. Compared to our base learning rate, reducing it by a factor of two led to flat loss. Increasing it by a factor of two led to divergence and NaNs.  



\section{Exporting Fuzzy Metaballs}\label{sec:marching}
\begin{figure}[tbh!]
     \centering
          \begin{subfigure}[t]{0.3\linewidth}
         \centering
         \includegraphics[width=1.0\linewidth]{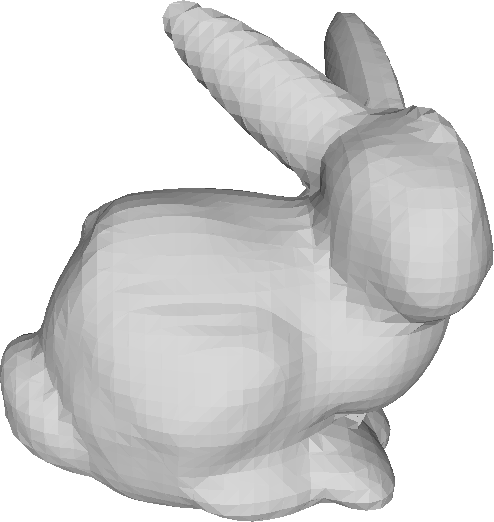}
     \end{subfigure}
         \hfill
          \begin{subfigure}[t]{0.3\linewidth}
         \centering
         \includegraphics[width=1.0\linewidth]{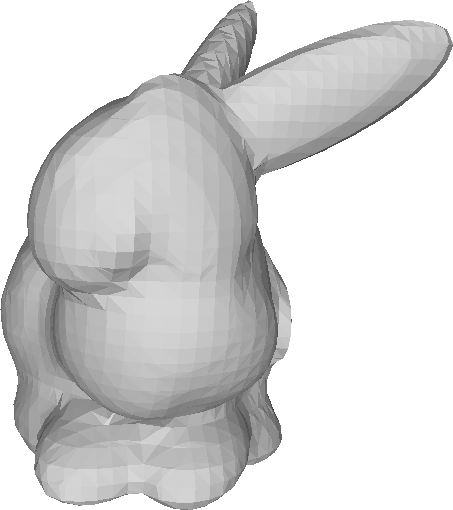}
     \end{subfigure}
              \hfill
          \begin{subfigure}[t]{0.3\linewidth}
         \centering
         \includegraphics[width=1.0\linewidth]{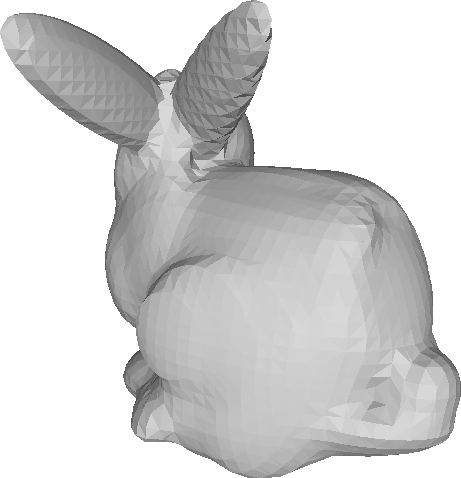}
     \end{subfigure}
    \caption{Mesh extracted from a 40 mixture fuzzy metaball using marching cubes  }
    \label{fig:march_cubes}
\end{figure}

We experiment with exporting fuzzy metaballs as a mesh by running marching cubes~\cite{lorensen1987marching}. To find an ideal isosurface level, we run optimization to ensure that the centroids of the voxels match the silhouettes over a sample set of views. This leads to results like those in ~\cref{fig:march_cubes}. 
\begin{figure}[tb]
     \centering
          \begin{subfigure}[t]{0.3\linewidth}
         \centering
         \includegraphics[width=1.0\linewidth]{marching/bunny_mesh00.png}
         \caption{40 component Fuzzy Metaballs (400 params)}
     \end{subfigure}
         \hfill
          \begin{subfigure}[t]{0.3\linewidth}
         \centering
         \includegraphics[width=1.0\linewidth]{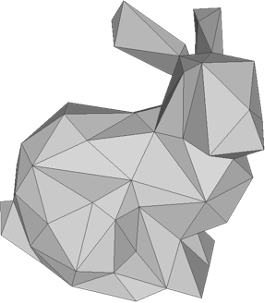}
         \caption{170 face, 85 vertex Mesh (810 params)}
     \end{subfigure}
              \hfill
          \begin{subfigure}[t]{0.3\linewidth}
         \centering
         \includegraphics[width=1.0\linewidth]{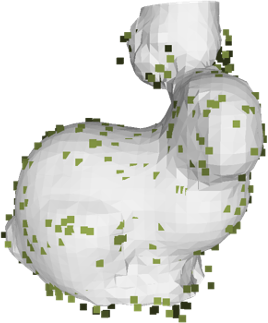}
         \caption{430 Points reconstructed~\cite{SGP:SGP06:061-070} (1290 params)}
     \end{subfigure}
    \caption{Equivalent representations visualized, per the experiments in the paper. }
    \label{fig:equiv_rep}
\end{figure}
\begin{figure}[tb]
     \centering
          \begin{subfigure}[t]{1.0\linewidth}
         \centering
         \includegraphics[width=1.0\linewidth]{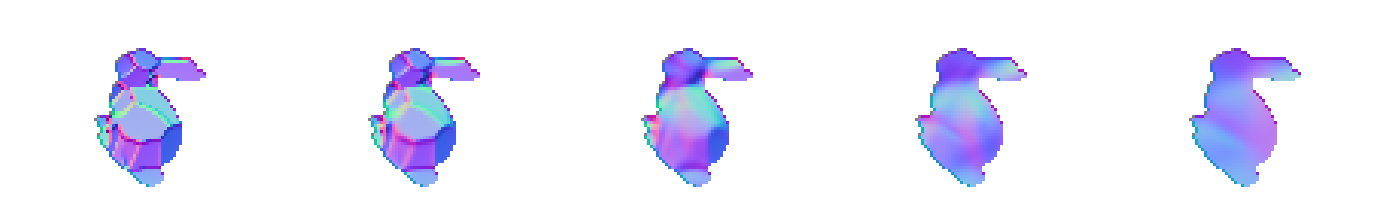}
         \caption{Visualizing normal maps while sweeping $\beta_1$ and $\beta_2$ demonstrates smoothing. }
     \end{subfigure}

              \vfill
          \begin{subfigure}[t]{1.0\linewidth}
         \centering
         \includegraphics[width=1.0\linewidth]{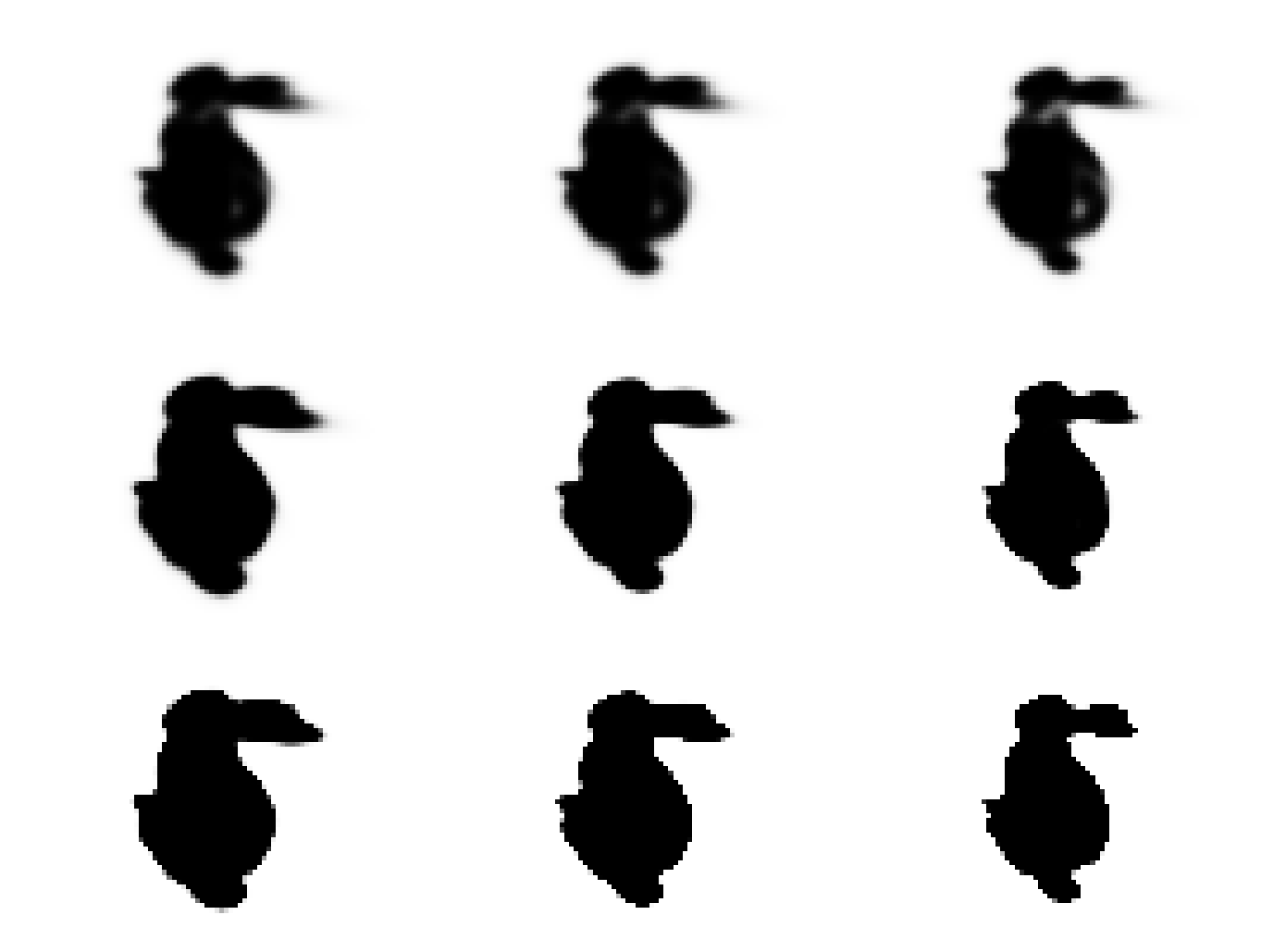}
         \caption{Sweeping $\beta_4$ and $\beta_5$ controls the sharpness and extent of the alpha masks. }
     \end{subfigure}
    \caption{Hyperparameter visualization }
    \label{fig:hyperparams}
\end{figure}

\section{Fuzzy Metaballs as Surface or Volume GMMs}\label{sec:supp_fit}

\begin{figure}[thb]
  \centering
   \includegraphics[width=1.0\linewidth]{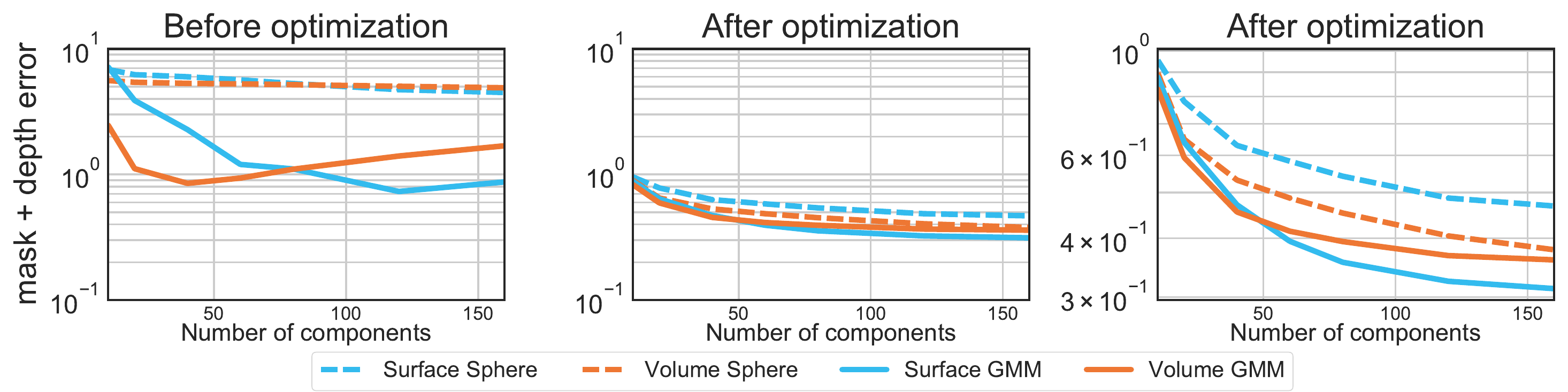}
    \caption{Optimizing Fuzzy Metaballs from different initializations.  }
    \label{fig:opt_starts}
\end{figure}

To understand what classical formulation best matches Fuzzy Metaballs, we try optimizing models with different initializations. We start with both sphere and EM-fit GMM initializations, with surface and volume versions of both. Quantitative results averaged across all 10 models are shown in \cref{fig:opt_starts}. Qualitative results for the \texttt{Yoga} model are shown in \cref{fig:vis_comp}. 

At low mixture numbers, Fuzzy Metaballs perform more like a volume GMM, while at high mixture numbers, surface GMMs work better. Often using a GMM as a FM model will produce reasonable results. We use constant hyperparameters from our 40 mixture tuning, and perhaps the out-of-the-box vGMM rendering could improve by adding proper scaling with component number. Finally, all initializations respond very well to optimization, and optimized sphere-initialized models always outperform the models fit with solely with EM. 

The Fuzzy Metaballs improve with more components across our entire range of testing. This suggests the asymptotic behavior seen in the \textbf{Comparing Representations} section is due to experimental factors of those experiments, and not the representation itself. This is somewhat expected as those experiments use the mesh representation as ground truth and all other formats are sampled.

Lastly, we can see that the fitting process produces no over-fitting as novel and training frames have identical behavior in \cref{fig:optimization}. 
\begin{figure}
  \centering
   \includegraphics[width=0.95\linewidth]{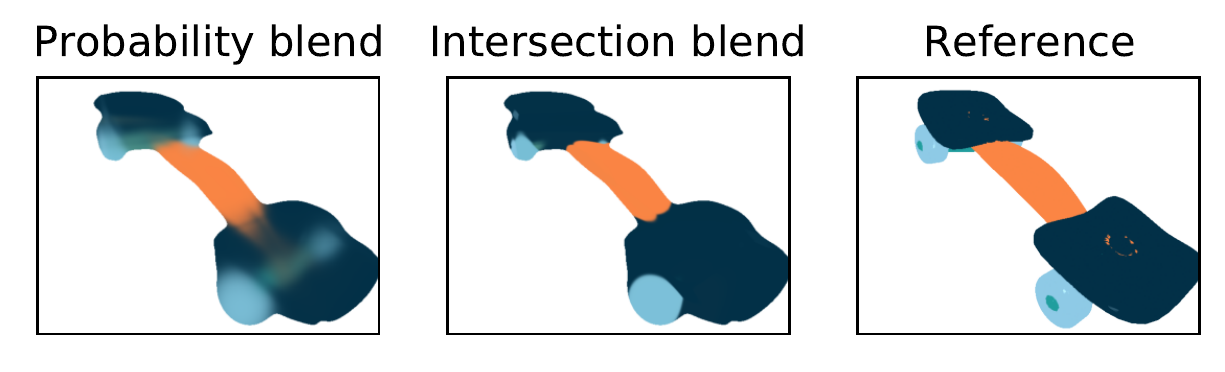}
    \caption{Rendering Fuzzy Metaball color images of a snakeboard~\cite{Kobilarov:2009:LGI} with two forms of blending: one behaves more like a volume where the wheels of the object can be seen, while the other behaves more like a surface with proper occlusion. Shown is a 40 component vGMM with a single color per component. Cartoon-like appearance is from exclusively using ambient lighting. }
    \label{fig:color_blend}
\end{figure}

\begin{figure}[ht]
  \centering
   \includegraphics[width=1.0\linewidth]{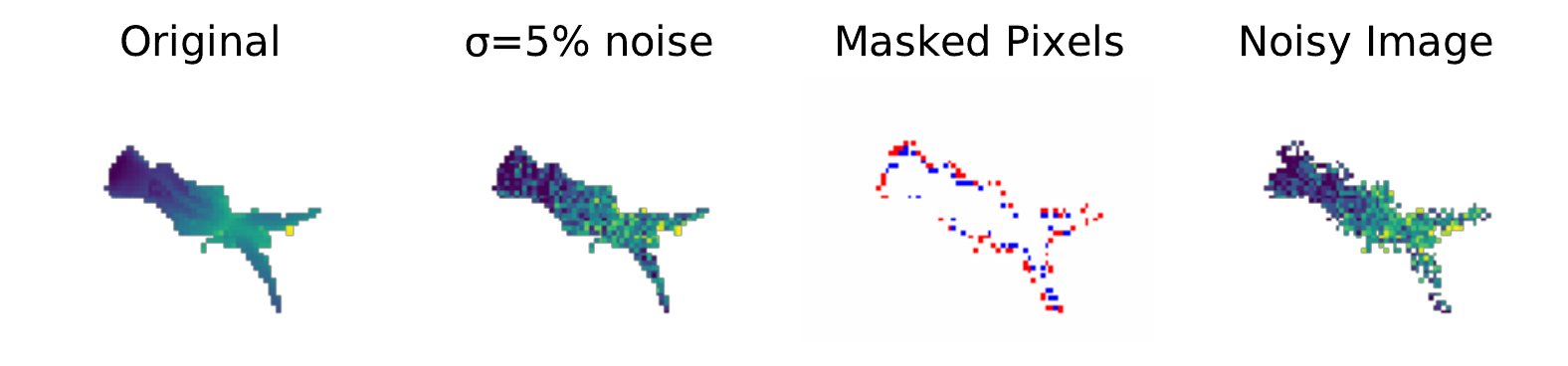}

    \caption{\textbf{Synthetic noise generation}. Gaussian noise is combined with perturbed silhouettes (red pixels are added, blue are removed). }
    \label{fig:noise_example}
\end{figure}
\begin{figure}[thb]
  \centering
   \includegraphics[width=0.7\linewidth]{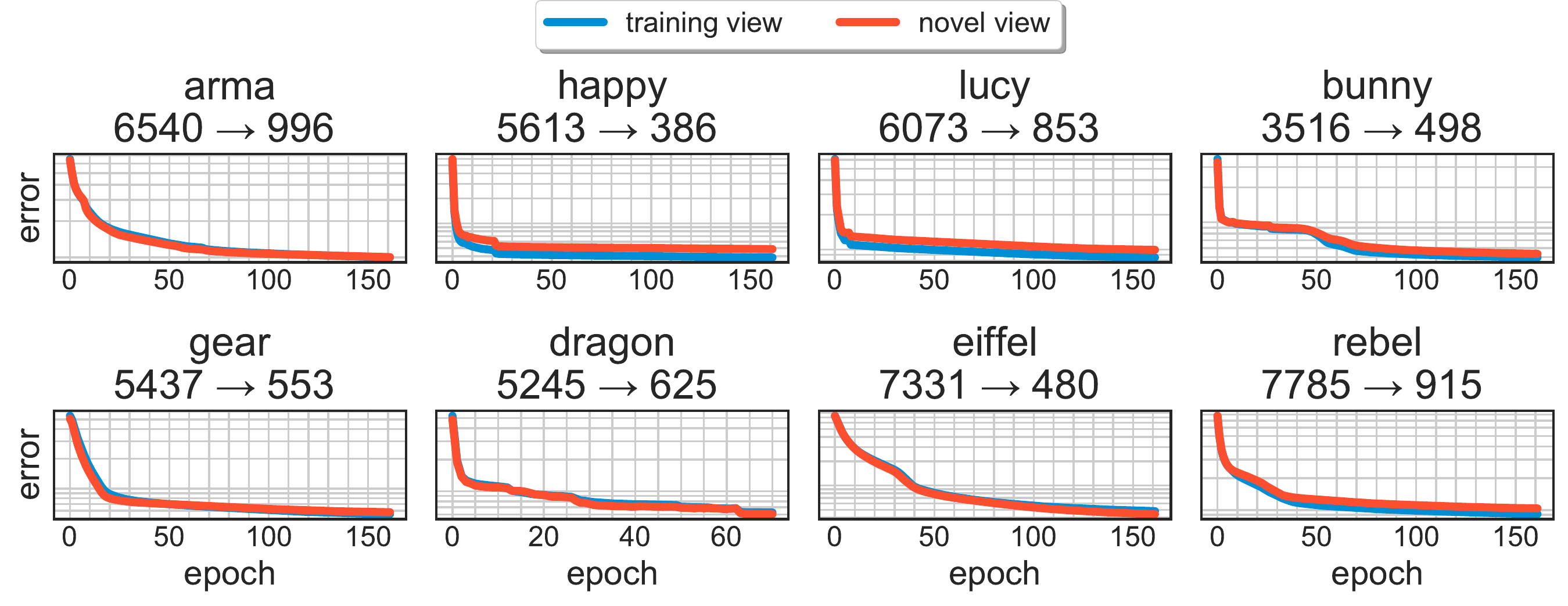}
    \caption{Optimizing Fuzzy Metaballs from a sphere to a shape. Losses are given for training frames and novel viewpoints, showing no significant difference. }
    \label{fig:optimization}
\end{figure}

\begin{figure}[t]
     \centering
          \begin{subfigure}[t]{1.0\linewidth}
         \centering
         \includegraphics[width=1.0\linewidth]{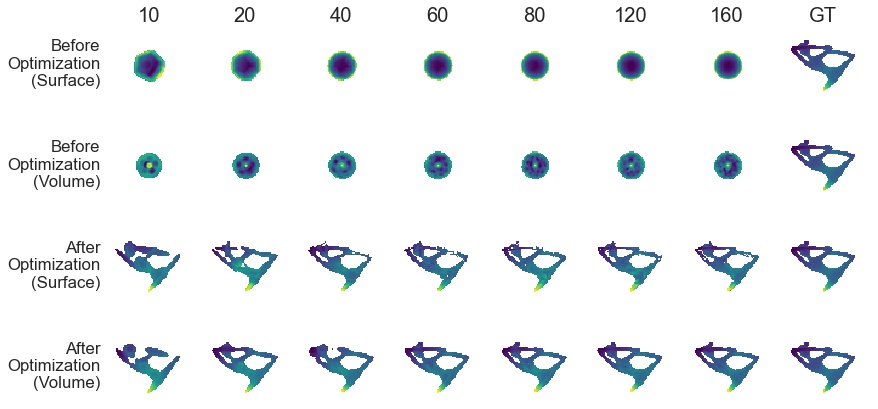}
    \caption{ Sphere Initialization }
    \label{fig:sphere_comp}
     \end{subfigure}
         \hfill
          \begin{subfigure}[t]{1.0\linewidth}
         \centering
         \includegraphics[width=1.0\linewidth]{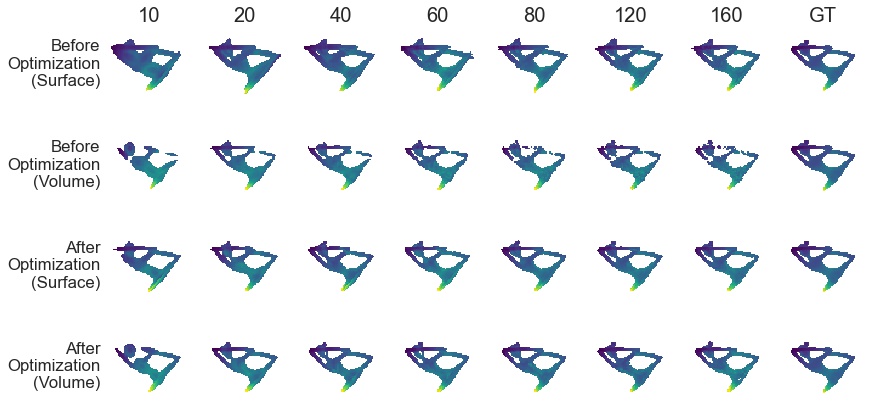}
    \caption{ GMM Initializations }
    \label{fig:gmm_comp}
     \end{subfigure}
    \caption{Visual examples of Fuzzy Metaballs at different component numbers, for different initializations, before and after optimization. All images are 60 by 80 pixels and show depth with color coding. Here, unlike the rest of the paper, colors are scaled for maximum contrast, not consistency between images. GT is the ground truth depth map from the mesh rendered by Blender.}
    \label{fig:vis_comp}
\end{figure}

\begin{figure}[tb]
     \centering
     \includegraphics[width=0.9\linewidth]{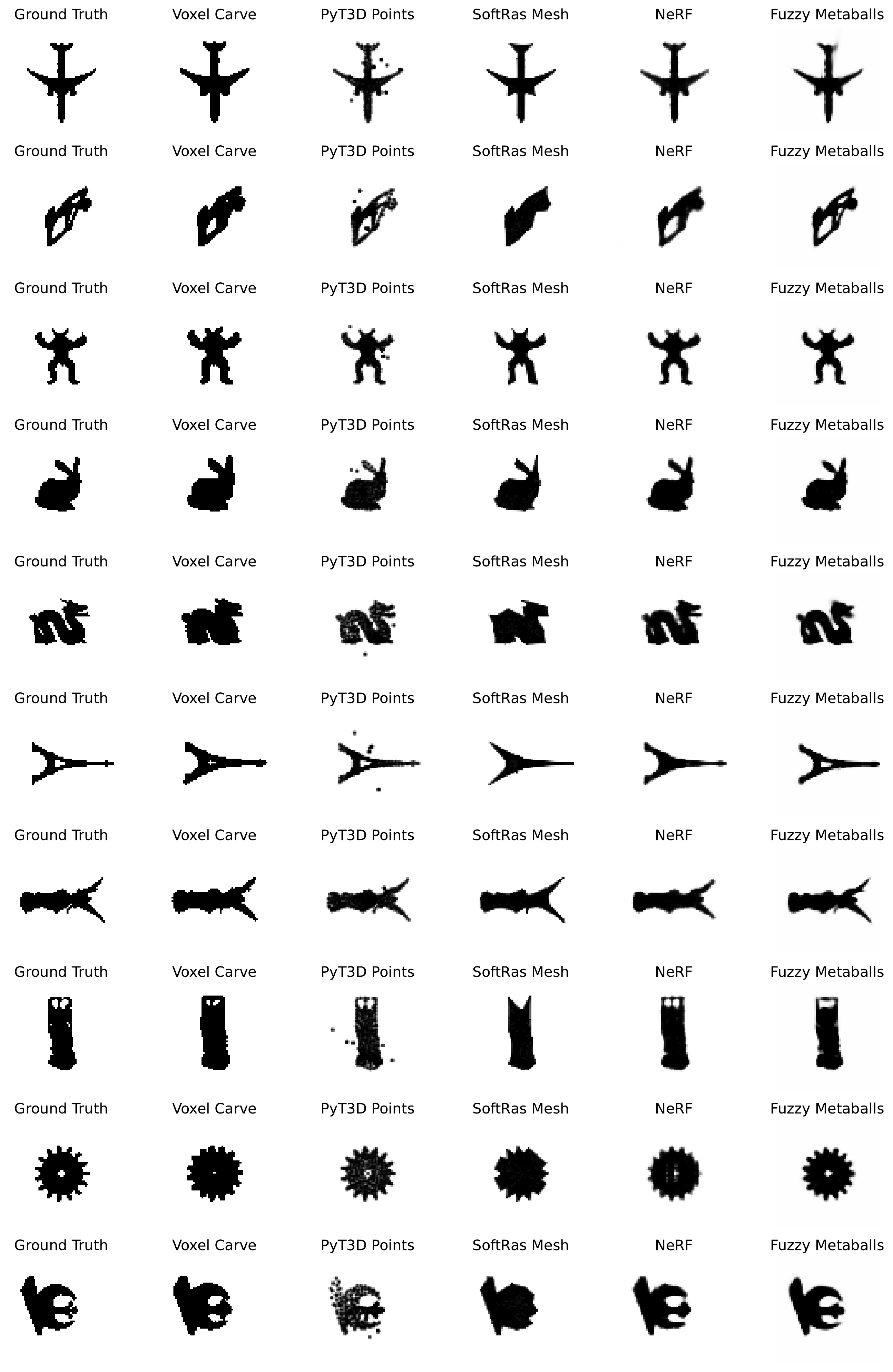}      
    \caption{\textbf{Shape from Silhouette Results}. The mesh-based representation cannot change genus from a deformed sphere into the eiffel tower. The point cloud method leaves spurious points. The classic Voxel Carving method is not that precise with $384^3$ volume but only 32 views of low resolution 64 x 64 images. }
    \label{fig:sill_holes_noisefree}
\end{figure}

\begin{figure}[tb]
     \centering
     \includegraphics[width=0.9\linewidth]{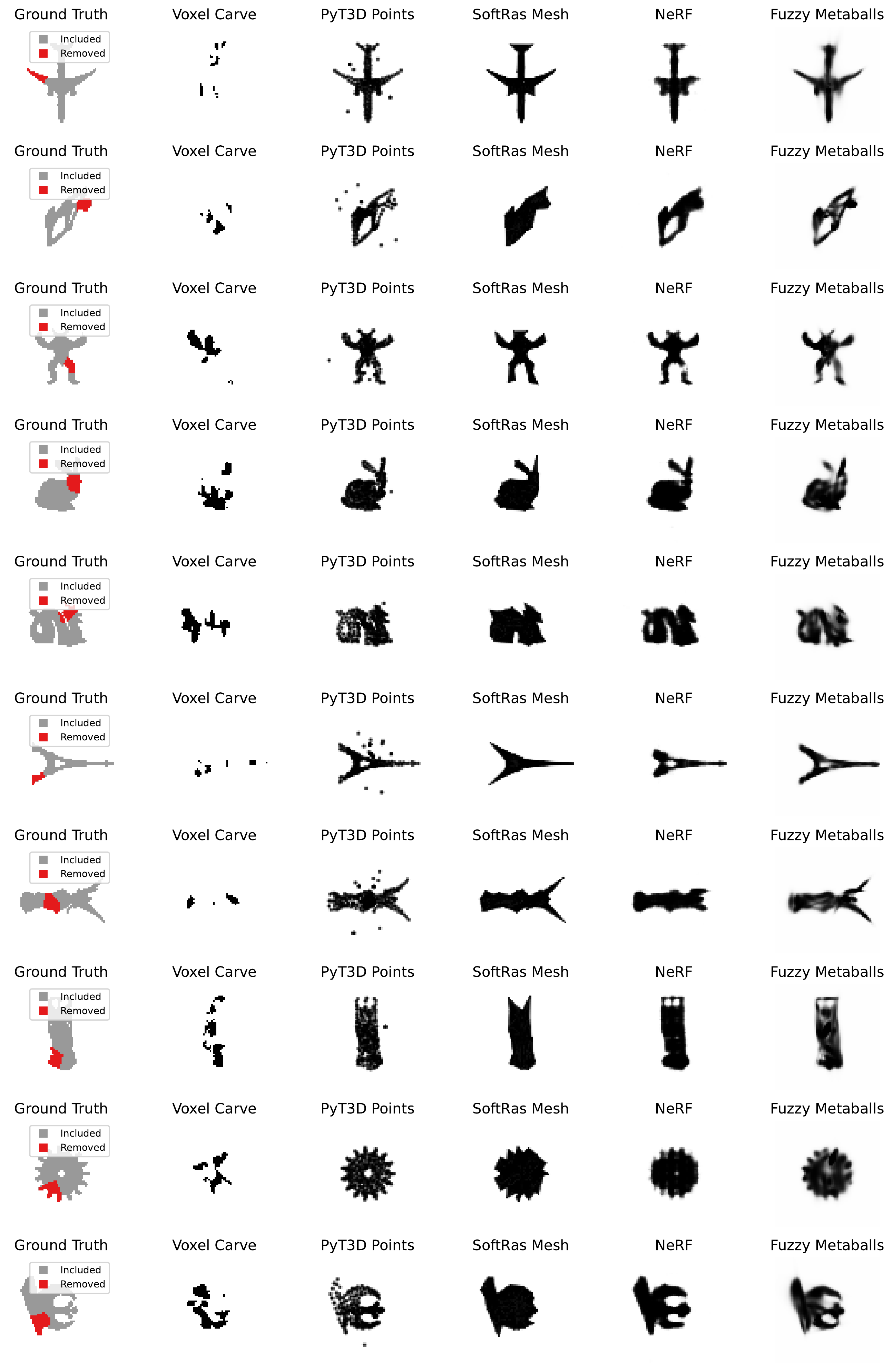}      
    \caption{\textbf{Shape from Silhouette Noisy Results} where 16 of the 32 input views had one eight of the silhouette under-segmented. }
    \label{fig:sill_holes_noisy}
\end{figure}
\end{document}